\documentclass[10pt,twocolumn,letterpaper]{article}

\usepackage{cvpr}
\usepackage{times}
\usepackage{epsfig}
\usepackage{graphicx}
\usepackage{amsmath}
\usepackage{amssymb}
\usepackage{amsthm}
\usepackage{nicefrac}
\usepackage{etoolbox}
\usepackage{bm}
\usepackage{mathtools}
\usepackage{algorithm}
\usepackage{algorithmic}
\usepackage{float}
\usepackage{caption}
\usepackage{subcaption}
\usepackage{xspace}
\usepackage{multirow}
\usepackage{array}
\usepackage{makecell}
\usepackage{stmaryrd}

\usepackage{booktabs}       

\usepackage[pagebackref=true,breaklinks=true,colorlinks,bookmarks=false]{hyperref}

\cvprfinalcopy 

\def\cvprPaperID{3264} 

\newtheorem{lem}{Lemma}

\newtheorem{prop}{Proposition}

\newtheorem{defi}{Definition}

\providecommand{\norm}[1]{\| #1 \|}

\newcommand{\R}{\mathbb{R}}
\newcommand{\N}{\mathbb{N}}

\newcommand{\np}{\textsc{np}}

\DeclareMathOperator*{\argmin}{arg\,min}

\ifcvprfinal\fi
\begin{document}

\title{Discrete-Continuous ADMM \\ for Transductive Inference in Higher-Order MRFs}

\author{\begin{minipage}{\textwidth}
    \centering
Emanuel Laude\,\textsuperscript{1} \quad
Jan-Hendrik Lange\,\textsuperscript{2} \quad
Jonas Sch\"upfer\,\textsuperscript{1}  \quad 
Csaba Domokos\,\textsuperscript{1}  \quad \\
Laura Leal-Taix{\'e}\,\textsuperscript{1}  \quad
Frank R. Schmidt\,\textsuperscript{1 3} \quad
Bjoern Andres\,\textsuperscript{2 3 4}  \quad
Daniel Cremers\,\textsuperscript{1}   \\[1ex]
\small
 \textsuperscript{1}\,Technical University of Munich \qquad \textsuperscript{2}\,Max Planck Institute for Informatics, Saarbr\"ucken \\
\textsuperscript{3}\,Bosch Center for Artificial Intelligence \qquad \textsuperscript{4}\,University of T\"ubingen
\end{minipage}
}

\maketitle

\begin{abstract}
This paper introduces a novel algorithm for transductive inference in higher-order MRFs, where the unary energies are parameterized by a variable classifier. The considered task is posed as a joint optimization problem in the continuous classifier parameters and the discrete label variables. In contrast to prior approaches such as convex relaxations, we propose an advantageous decoupling of the objective function into discrete and continuous subproblems and a novel, efficient optimization method related to ADMM. This approach preserves integrality of the discrete label variables and guarantees global convergence to a critical point. 
We demonstrate the advantages of our approach in several experiments including video object segmentation on the DAVIS data set and interactive image segmentation.
\end{abstract}

\section{Introduction}
Various problems in computer vision, computer graphics and machine learning can be formulated as MAP inference in a (possibly higher order) Markov random field (MRF)~\cite{lafferty2001conditional,koller09probabilistic,nowozin10structured,Kolmogorov2004,komodakis2007fast, delong2010,ladicky2014associative,paulsen2010markov}. The resulting optimization problem is defined over a hypergraph $(\mathcal{V},\mathcal{C})$ and a finite label set $\mathcal{L}$ as:
\begin{align} \label{eq:mrf}
\min_{y \in \mathcal{Y}} ~ \sum_{i\in \mathcal{V}} E_i(y_i) +\sum_{\substack{C \in \mathcal{C}\\ |C|>1}} E_{C}(y_C).
\end{align}
The optimization variable $y \in \mathcal{Y}:=\mathcal{L}^{|\mathcal{V}|}$ corresponds to a labeling of the vertices $\mathcal{V}$ and assigns a label $y_i \in \mathcal{L}$ to each vertex $i \in \mathcal{V}$. For convenience, we make a distinction between the singleton clique energies (unaries) $E_i(y_i)$ and the higher order energies $E_C(y_C)$, $|C|>1$.

\begin{figure}
\def\myHeight{6.45em}
\setlength{\tabcolsep}{1pt}
\begin{center}\begin{tabular}{cc}
\includegraphics[height=\myHeight]{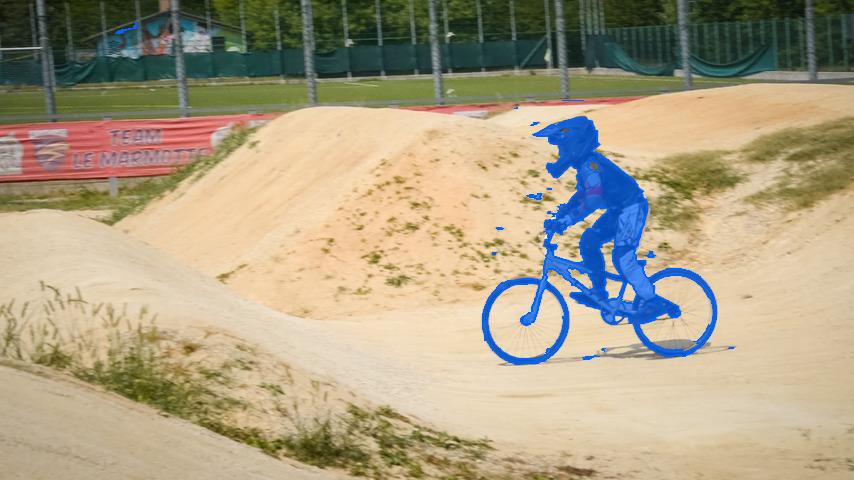} &
\includegraphics[height=\myHeight]{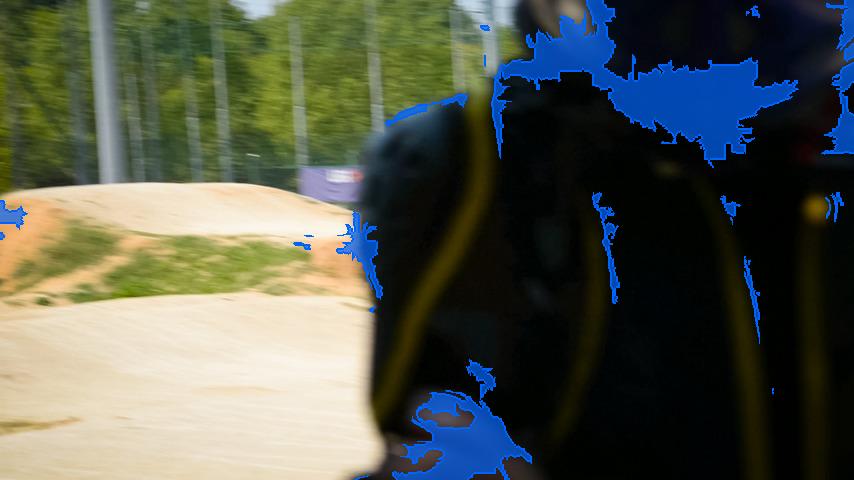} \\
\includegraphics[height=\myHeight]{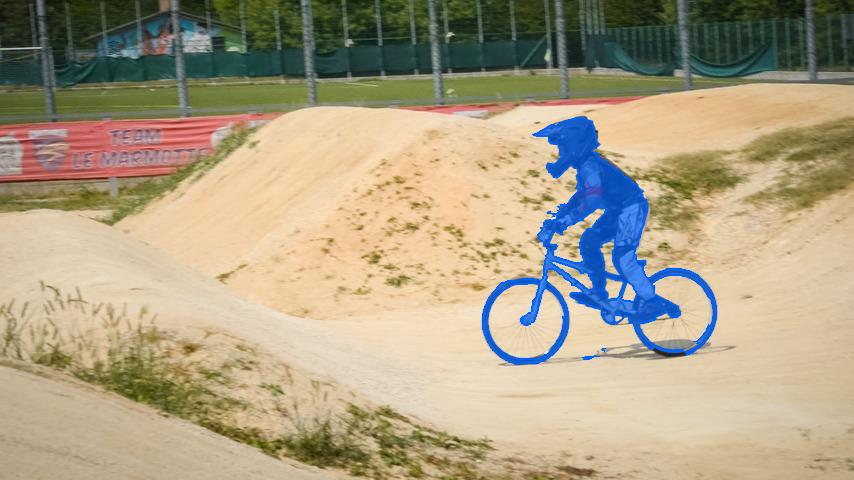} &
\includegraphics[height=\myHeight]{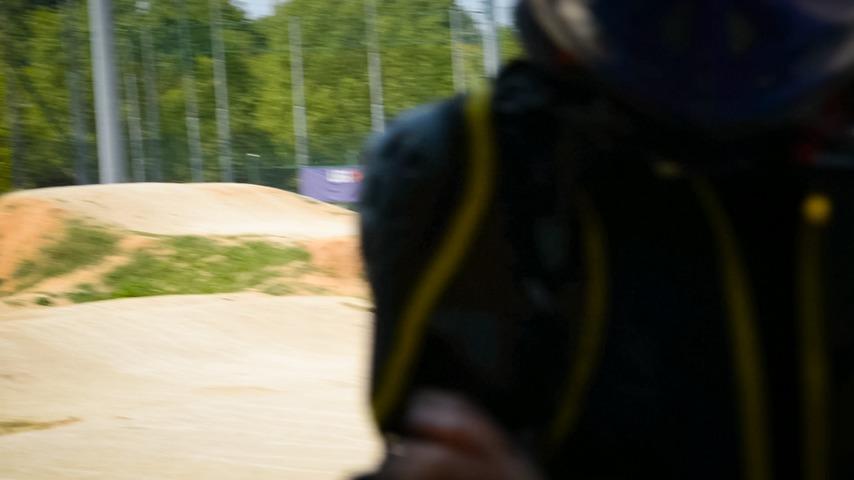} 
\end{tabular}\end{center}
\caption{A pixel-classifier trained to predict an object mask (\textbf{top row}) performs well when the distribution of object pixels in the training image is similar to the test image (\textbf{left}), but often fails if it is dissimilar (\textbf{right}). In a transductive inference approach (\textbf{bottom row}), we optimize jointly for the test labels and classifier parameters, which successfully prevents the hallucination of object pixels in difficult scenes, such as in the case of occlusion (\textbf{right}), cf.\ Sec.~\ref{sec:video-segmentation}.}
\label{fig:teaser}
\end{figure}

In computer vision tasks, where the image is interpreted as a (higher order) pixelgrid $(\mathcal{V}, \mathcal{C})$, the higher order potentials often correspond to priors favoring spatially smooth solutions.
In semantic image segmentation, for instance, inference in MRFs is widely used as a post-processing step to introduce spatial smoothness on the labeling $y$ \cite{chen:ICLR:2015}. In this sense, the overall task of semantic segmentation is subdivided into two tasks: First, a classifier, parameterized by $W$, is trained in a supervised fashion on a sufficiently large labeled training set, which assigns to each pixel in a test image a class (probability) score.
Second, to enforce spatial smoothness of the labeling of the test image, MAP-inference in an MRF is performed in a post-processing, in which the class (probability) scores are interpreted as the unary energies $E_i(y_i;W)$ for each $y_i$.
We argue that it is advantageous to merge such a two-step approach into a joint approach, as the training of the classifier profits from both, (i) the distribution of the unlabeled pixels in the color or feature space, and (ii) the available structural information about the unlabeled pixels, namely the spatial smoothness prior. Conversely, the segmentation will also profit from an improved classifier.


A joint formulation has two interpretations; on the one hand, it is a semi-supervised learning method that makes use of structural knowledge about the training data to learn a classifier. Such knowledge may take the form of higher-order clique energies $E_C$ \cite{wagstaff2001constrained,Basu-et-al-2008} on the labeling and acts as weak supervision in the training process.
This approach helps to mitigate the need for large amounts of annotated training data in typical modern machine learning applications.
We, on the other hand, focus on its interpretation as a \emph{transductive} inference method \cite{vapnik2013nature}, i.e.\ the approach to directly infer the labels of specific test data given specific training data, which we accomplish by incorporating a variable classifier in the inference process. Transductive inference stands in contrast to \emph{inductive} inference, which refers to first learning a general model from training data and subsequently applying the model to predict labels of a-priori unknown test data.
We show the benefits of using transductive inference for the tasks of video object segmentation (cf.\ Fig.~\ref{fig:teaser}) as well as scribble-based segmentation (cf.\ Fig.~\ref{fig:userInteraction}).

\subsection{Contributions}
We propose a general joint model that assumes the unaries not be fixed for inference in the MRF, but rather optimized jointly with the labeling.
Let for each $i\in \mathcal{V}$, $x_i \in \R^d$ denote the $d$-dimensional feature vector associated to the $i$th vertex and let $\varphi:\R^d \to \R^{d'}$ be a feature map.
Then, mathematically, such a task can be naturally formulated in terms of a bilevel optimization problem:
\begin{align}
\min_{\substack{y \in \mathcal{Y}, \\ W \in \R^{|\mathcal{L}| \times d'}}}&&& \sum_{i\in \mathcal{V}} \ell(y_i; W \varphi(x_i)) +\sum_{C \in \mathcal{C}} E_C(y_C) \label{eq:bilevel_model} \\
\text{subject to} &&& W = \argmin_{W \in \R^{|\mathcal{L}| \times d'}} \sum_{i\in \mathcal{V}} \ell(y_i; W  \varphi(x_i)) + g(W) \nonumber.
\end{align}
Here, the upper-level task is inference in a MRF with additional unaries $E_i(y_i; W):=\ell(y_i; W \varphi(x_i))$, parameterized by linear classifier weights $W \in \R^{|\mathcal{L}| \times d'}$, and a loss function $\ell \colon \mathcal{L} \times \R^{|\mathcal{L}|} \to \R$. The lower-level task associates to each given set of labels $y$ the optimal parameters $W$. For instance, if $\ell$ is the hinge loss and $g(W) = \norm{W}_F^2$, then the lower-level optimization problem amounts to the training of a classical SVM. Note that in a semi-supervised learning context it might be more convenient to swap upper and lower level tasks, since the primary interest is the estimation of the classifier that minimizes the generalization error and not the inferred labeling. However, mathematically, both viewpoints are equivalent.

The model \eqref{eq:bilevel_model} suggests a simple alternating optimization scheme as in Lloyd's algorithm~\cite{lloyd1982least} for $k$-means to compute a local optimum. However, such an approach has two major drawbacks: (i) The lower-lever subproblems are expensive, which is prohibitive for large scale applications. (ii) The optimization is prone to poor local optima and therefore sensitive to initialization \cite{zhang2009maximum}.
Motivated by the good practical performance of the alternating direction method of multipliers (ADMM) in nonconvex optimization, we propose to generalize vanilla ADMM (commonly applied in continuous optimization) to discrete-continuous problems of the form \eqref{eq:bilevel_model}, while preserving integrality of the discrete variables. Since our method serves as a general algorithmic framework to tackle such problems, it is also relevant to semi-supervised and transductive learning in a broader sense.

The main contributions of this work can be summarized as follows:
\begin{itemize}
\item We devise a decomposition of the model into simple, purely discrete and purely continuous subproblems within the framework of proximal splitting. The subproblems can be solved in a distributed fashion.
\item We devise a tailored ADMM-inspired algorithm, \emph{discrete-continuous ADMM}, to compute a local optimum of \eqref{eq:bilevel_model}. In contrast to vanilla ADMM, our algorithm allows us to obtain sub-optimal solutions of the MAP inference problem so that also computationally more challenging MRFs can be considered.
\item We generalize the convergence of nonconvex vanilla ADMM to the presented inexact discrete-continuous ADMM.
\item In diverse experiments we demonstrate the relevance and generality of our model and the efficiency of our method: In contrast to standard $k$-means, our model integrates well with deep features. In contrast to a tailored SDP relaxation approach for transductive logistic regression, our method produces more consistent results, while being more efficient in terms of both runtime and memory consumption.
\end{itemize}

\subsection{Related work}
To improve image segmentation results it is common practice to treat the unary terms $E_i$ as additional variables in the optimization \cite{blake1987visual,zhu1996region,chan2001active,rother2004grabcut,tang2015secrets,tang2016normalized, TrajkovskaSAP17}. More recently, \cite{tang2015secrets,tang2016normalized} revealed the equivalence of $k$-means clustering with pairwise constraints \cite{wagstaff2001constrained,Basu-et-al-2008} and the Chan-Vese \cite{chan2001active} approach, where the average foreground and background intensities (corresponding to the centroids in $k$-means) are not assumed to be fixed, but are rather treated as additional variables. 
The goal of this approach is to jointly cluster the pixels in the color space and regularize the cluster-assignment (the segmentation) in the image space. The clustering viewpoint suggests the application of the ``kernel trick'', which allows us to separate more complicated, possibly nonlinearly deformed color clusters \cite{wagstaff2001constrained,Basu-et-al-2008,tang2015secrets,tang2016normalized}. Experimentally, it has been shown that this approach integrates well with color or even depth pixel features \cite{tang2015secrets,tang2016normalized}.
Due to the enormous success of deep convolutional neural networks on computer vision tasks, it is tempting to replace the color features by more sophisticated deep features that are capable of compactly representing complicated semantic information \cite{alexnet:NIPS:2012,chen:ICLR:2015}.
However, high-dimensional deep features are in general not ``$k$-means friendly'' \cite{YangFSH17} and without further preprocessing of the features as in \cite{YangFSH17} the plain Chan-Vese $k$-means approach does not generalize very well to deep features, despite of (almost) linear separability of the data.
Since deep neural network classifiers can be viewed as a linear model on top of a deep feature extractor, we propose to alter the approaches from related work by using a (multiclass) SVM or a (multinomial) logistic regression model along with deep features.
Under the absence of general higher order terms (i.e.\ $E_C=0$, for all $C \in \mathcal{C}$) our model \eqref{eq:bilevel_model} is closely related to transductive SVMs \cite{vapnik2013nature,joachims1999transductive,bennett1999semi} and transductive logistic regression \cite{DBLP:conf/icml/JoulinB12}. 
In such a setting, optimization schemes that alternately optimize w.r.t.\ to labels and model parameters as in Lloyd's algorithm~\cite{lloyd1982least} are ineffective \cite{zhang2009maximum}. Other approaches, such as SDP relaxations are computationally expensive \cite{DBLP:conf/icml/JoulinB12}.

Instead, we propose an algorithm related to ADMM, which has recently been successfully applied to many nonconvex continuous optimization problems \cite{chartrand2013,ozolicnvs2013compressed,storath2014jump,miksik2014distributed,lai2014splitting}.
ADMM appears similar in form to message passing and subgradient descent schemes applied to the Lagrangian dual problem (dual decomposition) \cite{komodakis2007mrf,DBLP:journals/ftml/BoydPCPE11,martins2011augmented,pmlr-v31-zach13a,swoboda-2017}. The latter is a Lagrangian relaxation approach, so that in difficult nonconvex cases the linear equality constraints may remain violated in the limit \cite{komodakis2007mrf}. In contrast, ADMM attempts to solve the problem exactly and enforce the linear equality constraints strictly via additional quadratic penalty terms.
In order to make mixed discrete-continuous problems such as \eqref{eq:bilevel_model} amenable to ADMM, related approaches often relax the discrete variable and perform rounding operations \cite{huang2016consensus,takapoui2017simple}. In contrast, we propose a generalization of vanilla ADMM that preserves the integrality of the label variable and admits a theoretical convergence guarantee under affordable conditions. 
In the traditional convex and continuous setting, ADMM \cite{glowinski1975approximation,gabay1976dual} converges under mild conditions \cite{gabay1983chapter,eckstein1992douglas}. For more restrictive nonconvex problems, its convergence has only been established recently \cite{hong2016convergence,li2015global}. In this case, however, the required assumptions are fairly strong.
\section{Discrete-Continuous ADMM}
The coupling of the discrete labeling variable $y$ and the continuous variable $W$ renders problem \eqref{eq:bilevel_model} hard to solve. This is not surprising since the related $k$-means clustering problem is known to be $\np$-hard. A common approach is to compute a local optimum by a simple discrete-continuous coordinate descent approach as in Lloyd's algorithm~\cite{lloyd1982least}. Instead, we propose an advantageous decoupling into purely discrete and purely continuous subproblems, which allows us to compute a local optimum by updating the continuous and discrete variables jointly and efficiently. 

\subsection{Variable decoupling via ADMM}
To this end, we employ a change of representation to make the proposed problem amenable to the ``kernel trick''. Note that, for any fixed labeling $y$, the lower-level task in \eqref{eq:bilevel_model} amounts to supervised SVM training (resp.\ supervised logistic regression). Thus, we can apply the representer theorem \cite{scholkopf2001generalized}: Let $\Phi(X)$ be the feature matrix for a (possibly infinte-dimensional) matrix feature map $\Phi:\R^{d \times |\mathcal{V}|} \to \R^{d' \times |\mathcal{V}|}$ and let 
\begin{align}
g(W) = h(\| W\|_F),
\end{align}
for $h \colon [0,\infty) \to \R$ strictly monotonically increasing. Then, the weights $W^\top =\Phi(X)\alpha$ can be substituted via their representation $\alpha\in \R^{|\mathcal{V}| \times |\mathcal{L}|}$ in terms of the features.
More precisely, we replace the scalar products $W \varphi(x_i)$, up to transposition, by $K_i \alpha=(W\varphi(x_i))^\top$ where $K:=\Phi(X)^\top \Phi(X)$ denotes the Gram or kernel matrix.

For $f:\R^{|\mathcal{V}| \times |\mathcal{L}|} \to \R$, being defined as
\begin{align} \label{eq:regularizer}
f(\alpha):=h( \|\Phi(X) \alpha\|_F),
\end{align}
this substitution leaves us with the following equivalent mixed integer nonlinear program formulation of \eqref{eq:bilevel_model}:
\begin{equation}\label{eq:onelevel_model}
\begin{aligned}
\underset{\substack{y \in \mathcal{Y}, \\ \alpha \in \R^{|\mathcal{V}| \times |\mathcal{L}|}}}{\min}\sum_{i\in \mathcal{V}} \ell(y_i; K_i \alpha) + f(\alpha) +\sum_{C \in \mathcal{C}} E_C(y).
\end{aligned}
\end{equation}
In order to decompose problem \eqref{eq:onelevel_model} into simple subproblems associated with each $i \in \mathcal{V}$, we introduce auxiliary variables $\beta_i = K_i \alpha$, which yields
\begin{equation} \label{eq:model_consensus}
\begin{aligned}
\underset{\substack{y \in \mathcal{Y}, \\ \alpha,\beta \in \R^{|\mathcal{V}| \times |\mathcal{L}|}}}{\min} &&&\sum_{i\in \mathcal{V}} \ell(y_i; \beta_i) + f( \alpha) +\sum_{C \in \mathcal{C}} E_C(y) \\
\text{subject to}  & &&  K \alpha = \beta.
\end{aligned}
\end{equation}
Note that the objective of \eqref{eq:model_consensus} is a separable function over the $\beta_i$. This suggests to relax the linear constraint $K\alpha = \beta$ and consider the equivalent saddle point problem:
\begin{equation} \label{eq:model_saddle_point}
\begin{aligned}
\underset{\substack{y \in \mathcal{Y}, \\ \alpha,\beta \in \R^{|\mathcal{V}| \times |\mathcal{L}|}}}{\min} \max_{\lambda\in \R^{|\mathcal{V}| \times |\mathcal{L}|}} & \mathfrak{L}_\rho(\alpha, \beta, \lambda, y),
\end{aligned}
\end{equation}
where $\lambda \in \R^{|\mathcal{V}| \times |\mathcal{L}|}$ are the Lagrange multipliers corresponding to $K\alpha = \beta$ and $\mathfrak{L}_\rho$ denotes the ``discrete-continuous'' augmented Lagrangian, that for some penalty parameter $\rho>0$ is defined as
\begin{equation} \label{eq:dc_augm_lagr}
\begin{aligned}
&\mathfrak{L}_\rho(\alpha, \beta, \lambda, y):=\sum_{i\in \mathcal{V}} \ell(y_i; \beta_i) + f( \alpha) \\
&\quad +\sum_{C \in \mathcal{C}} E_C(y)+\langle \lambda, K \alpha - \beta \rangle + \frac{\rho}{2}\|K \alpha - \beta\|_F^2.
\end{aligned}
\end{equation}
We show in Sec.~\ref{sec:subproblems} that, for fixed $\lambda$ and $\alpha$, the function $\mathfrak{L}_\rho(\alpha, \cdot, \lambda, \cdot)$ can be minimized (not necessarily to global optimality) efficiently and jointly over $y$ and $\beta$.
This central observation and the good practical performance of ADMM in nonconvex optimization motivates the following generalization to discrete-continuous problems of the form \eqref{eq:model_saddle_point}.

We propose an algorithm that, similar to continuous ADMM, updates the discrete-continuous variable-pair $(\beta^{t+1},y^{t+1})$ via joint (and possibly suboptimal) minimization of $\mathfrak{L}_\rho(\alpha^{t}, \cdot, \lambda^t, \cdot)$. Subsequently, it updates $\alpha^{t+1}$ via minimization of $\mathfrak{L}_\rho(\cdot, \beta^{t+1}, \lambda^t, y^{t+1})$ and the Lagrange multiplier $\lambda^t$ by performing one iteration of gradient ascent on $\mathfrak{L}_\rho(\alpha^{t+1}, \beta^{t+1}, \cdot, y^{t+1})$ with step size $\rho>0$. In summary, the update steps at iteration $t$ are given as
\begin{align}
(\beta^{t+1}, y^{t+1}) &= \textstyle\argmin_{\beta, y} \mathfrak{L}_\rho(\alpha^{t}, \beta, \lambda^t, y), \label{eq:update_discrete_contiuous} \\
\alpha^{t+1} &= \textstyle\argmin_{\alpha} \mathfrak{L}_\rho(\alpha, \beta^{t+1}, \lambda^t, y^{t+1}), \label{eq:update_alpha} \\
\lambda^{t+1} &= \lambda^t + \rho(K \alpha^{t+1} - \beta^{t+1})  \label{eq:update_dual}.
\end{align}
In practice, we choose the step size adaptively, as this often leads to better solutions in terms of objective value: For finitely many iterations, the penalty parameter $\rho$ is increased according to the schedule $\rho_{t+1} = \min \; \{ \rho_{\max}, \tau \rho_t \}$ with $\tau>1$ and some $\rho_{\max}>0$ that guarantees theoretical convergence of the algorithm (cf.\ Sec.~\ref{sec:convergence}).


\subsection{Distributed solution of the subproblems} \label{sec:subproblems}
In this section, we describe the implementation of update steps \eqref{eq:update_discrete_contiuous}--\eqref{eq:update_dual} in our algorithm. In principle, \eqref{eq:update_discrete_contiuous} could be solved by minimization over $\beta$ for every feasible labeling $y \in \mathcal{Y}$.
Obviously, this is not a viable approach, as it implies performing exhaustive search over the set $\mathcal{Y}$, which has size $\lvert \mathcal{L}\rvert^{\lvert \mathcal{V} \rvert}$. Instead, we pursue the following more efficient strategy.

\paragraph{Solution via lookup-tables.}
Assume first the absence of any higher order energies, i.e.\ $E_C = 0$, for all $C \in \mathcal{C}$.
Then, since $\mathfrak{L}_\rho(\alpha^{t}, \beta, \lambda^t, y)$ is separable w.r.t.\ $\beta_i$ and $y_i$, we can decompose problem \eqref{eq:update_discrete_contiuous} into $|\mathcal{V}|$ independent problems of the form
\begin{align} \label{eq:subproblem}
\argmin_{\beta_i, y_i} ~\underbrace{\ell(y_i; \beta_i) +\frac{\rho}{2} \|\beta_i - K_i \alpha^t - \nicefrac{\lambda_i^t}{\rho} \|_F^2}_{\psi_i(\beta_i, y_i;\alpha^t, \lambda_i^t)},
\end{align}
which can thus be solved in parallel.
In the presence of higher-order energies, however, the problems \eqref{eq:subproblem} are not completely independent, because the variables $y_i$ are coupled via the energies $E_C$ in which they appear. In this case, we first solve \eqref{eq:subproblem} w.r.t.\ only the continuous variables $\beta_i$ for \emph{every} possible label $y_i \in \mathcal{L}$ and store the results in a lookup-table $(u^{t+1}, B^{t+1})$.

Precisely, for each $1\leq i \leq |\mathcal{V}|$ and each $y_i \in \mathcal{L}$ we create an entry $(u_{i,y_i}^{t+1}, B_{i,y_i}^{t+1})$ according to
\begin{equation} \label{eq:lookup_table}
\begin{aligned}
&B_{i,y_i}^{t+1}:= \argmin_{\beta_{i}} \psi_i(\beta_i; y_i, \alpha^t, \lambda_i^t), \\
&u_{i,y_i}^{t+1} := \min_{\beta_{i}} \psi_i(\beta_i; y_i, \alpha^t, \lambda_i^t).
\end{aligned}
\end{equation}

In a second step, we determine the discrete variable update $y^{t+1}$ as the (possibly suboptimal) solution of the MRF
\begin{align} \label{eq:subproblem_mrf}
y^{t+1}=\argmin_{y \in \mathcal{Y}}\sum_{i\in \mathcal{V}} u_{i,{y_i}}^{t+1} + \sum_{C\in \mathcal{C}} E_C(y).
\end{align}
Afterwards, the continuous variable updates $\beta_i^{t+1}$ can be read off from the solution of \eqref{eq:lookup_table} via 
\begin{align} \label{eq:update_consensus}
\beta_i^{t+1}=B_{i, y_i^{t+1}}^{t+1}.
\end{align}

Note that there is an abundance of algorithms available to tackle problems of the form \eqref{eq:subproblem_mrf} such as graph cuts for binary submodular MRFs \cite{Kolmogorov2004}, move making and message passing algorithms \cite{delong2010,kolmogorov2015new}, primal-dual algorithms \cite{komodakis2007fast,fix2014primal} and more. For an overview, see also \cite{kappes2013}.

The matrix $u$ specifies the unary energies in problem \eqref{eq:subproblem_mrf} that pushes the MRF to attain a labeling which corresponds to a more suitable classifier. The latter is determined by a tradeoff between minimizing the distance of $\beta_i$ to the current consensus parameters $K_i \alpha^t + \nicefrac{\lambda_i^t}{\rho}$ and minimizing the loss term corresponding to sample $i$.

In case of suboptimality of $y^{t+1}$ we require that $y^{t+1}$, for some $\delta \geq 0$, satisfies a (sufficient) descent condition
\begin{align} \label{eq:suff_descent_discrete}
\mathfrak{L}_\rho(\alpha^t, \beta^{t+1}, \lambda^t, y^{t+1})  - \mathfrak{L}_\rho(\alpha^t, B_{:, y^t}^{t+1}, \lambda^t, y^t) \leq -\delta.
\end{align}
If this condition is violated, then we keep the previous iterate $y^{t+1}=y^t$.
Under condition \eqref{eq:suff_descent_discrete}, the overall convergence of our algorithm is guaranteed (cf.\ Prop.~\ref{prop:convergence_nonconvex} and Prop.~\ref{prop:convergence_convex} in Sec.~\ref{sec:convergence}). We summarize our method in Alg.~\ref{alg:dc_admm}.

\setlength{\textfloatsep}{10pt}
\begin{algorithm}[tbp]
\caption{Discrete-Continuous ADMM}
\label{alg:dc_admm}
\begin{algorithmic}[1]
\REQUIRE initialize $\alpha^0, \lambda^0, \rho_0 > 0, \tau > 1$, $\rho_{\max}$ as in \eqref{eq:penalty_condition} 
\WHILE{(not converged)}
\STATE Compute lookup-table $(u^{t+1},B^{t+1})$:
   \FORALL{$i\in\{1,\hdots,|\mathcal{V}|\}$ \AND $y_j \in \mathcal{L}$}
	\STATE In parallel update $(u_{i,y_j}^{t+1}, B_{i,y_j}^{t+1})$ as in \eqref{eq:lookup_table}.
   \ENDFOR
\STATE Update $y^{t+1}$ as in \eqref{eq:subproblem_mrf}.
\IF{$y^{t+1}$ violates condition \eqref{eq:suff_descent_discrete}}
\STATE $y^{t+1} \gets y^{t}$.
\ENDIF
\STATE $\beta_i^{t+1}\gets B_{i, y_i^{t+1}}^{t+1}$
\STATE Perform updates, as in \eqref{eq:update_alpha},\eqref{eq:update_dual}.
\IF{$\rho$ violates condition \eqref{eq:penalty_condition}}
\STATE $\rho_{t+1} \gets \min \; \{ \rho_{\max}, \tau \rho_t \}$.
\ENDIF
\STATE $t \gets t+1$
\ENDWHILE
\end{algorithmic}
\end{algorithm}

Note that if the discrete subproblem \eqref{eq:subproblem_mrf} is solved to global optimality, our method specializes to classical nonconvex ADMM applied to a purely continuous problem $\min_{\alpha} E(K\alpha) + f(\alpha)$. The function  $E(\beta) = \min_{y\in \mathcal{Y}} \sum_{i\in \mathcal{V}} \ell(y_i; \beta) + \sum_{C \in \mathcal{C}} E_C(y)$ encapsulates the minimization over the discrete labelings $y$. This results in a pointwise minimum over exponentially many functions.

\paragraph{Distributed optimization.}
Distributed optimization is considered one of the main advantages of ADMM in supervised learning \cite{Forero:2010,DBLP:journals/ftml/BoydPCPE11}.
In our method, the $(\beta,y)$ update requires the solution of only $\lvert \mathcal{L} \rvert \cdot |\mathcal{V}|$ many (instead of $\lvert \mathcal{L}\rvert^{\lvert \mathcal{V} \rvert}$ for the naive approach) independent and small-scale continuous minimization problems of the form \eqref{eq:lookup_table} and one additional discrete problem \eqref{eq:subproblem_mrf}. This suggests the distributed solution of the subproblems \eqref{eq:lookup_table}, for instance on a GPU. Subsequently, the optimization of the MRF \eqref{eq:subproblem_mrf} and the update of the consensus variable is carried out after gathering the solutions of the subproblems. Since \eqref{eq:subproblem_mrf} need not be solved to optimality, the MRF solver may be stopped early to speed up computation. This is particularly useful if a primal-dual algorithm for solving the LP-relaxation is used \cite{komodakis2007fast,fix2014primal}.

\paragraph{Exploit duality.}
If the loss terms $\ell(y_i; \cdot)$ are convex and lower semicontinuous, then the independent subproblems \eqref{eq:lookup_table} can be solved efficiently via duality as follows. For all loss functions we consider, it is convenient to solve the dual problem as it scales linearly with the number of training samples (which is equal to one in our case). For the Crammer and Singer multiclass SVM loss \cite{crammer2001algorithmic}, for instance, there exists an efficient variable fixing algorithm \cite{Kiwiel2008} for solving the dual problem. For the softmax loss the dual problem reduces to a one-dimensional nonlinear equation via the Lambert-$W$ function \cite{Lapin:2016} and may be solved by performing a few iterations of Newton's or Halley's method. For the special case of the one-vs.-all hinge loss, \eqref{eq:lookup_table} can be solved in closed form. In any case, each subproblem involves only a small number of instructions, which is important for a GPU-based implementation.

\subsection{Consensus update}
For a quadratic regularizer $h(x)=\nu x^2$, where $\nu$ is the regularization parameter, the update step \eqref{eq:update_alpha} is equivalent to
\begin{align}
\alpha^{t+1}=\argmin_{\alpha}~ \nu\, \langle \alpha, K \alpha \rangle + \frac{\rho}{2} \|K \alpha - \beta^{t+1} + \nicefrac{\lambda^t}{\rho} \|_F^2.
\end{align}
This is a quadratic problem that can be solved via a normal equation using either a cached eigenvalue decomposition of the kernel matrix, or an iterative algorithm such as conjugate gradient (CG). The latter is preferred for large scale applications, as each CG iteration involves a kernel-matrix-vector multiplication $K v$. For the linear kernel $K=X^\top X$, this guarantees efficiency of our method, since $K$ does not have to be stored explicitly. For general kernels such as the RBF kernel, a low rank approximation to the kernel matrix $K\approx GG^\top$, for some $G \in \R^{|\mathcal{V}| \times l}$ with $l \ll |\mathcal{V}|$ can be obtained for instance via the Nystr\"om method \cite{nystrom1930praktische,williams2001using} or random features \cite{rahimi2008random}. Furthermore, in practice, often only a small number of conjugate gradient iterations are necessary. 

\section{Convergence analysis}
\label{sec:convergence}
In this section, we provide a complete convergence analysis of the proposed algorithm. To this end, we make the following assumptions:
\begin{itemize}
\item The function $f$ is $L$-smooth, $m$-semiconvex and lower-bounded, i.e.\ $f$ is differentiable and $\nabla f$ is Lipschitz-continuous with modulus $L$ and there exists $m>0$ sufficiently large so that $f+\frac{m}{2}\|\cdot\|_F^2$ is convex.
\item For all $y_i \in \mathcal{L}$, $\ell(y_i; \cdot)$ is lower-bounded.
\item The kernel matrix $K \in \R^{|\mathcal{V}| \times |\mathcal{V}|}$ is surjective, i.e.\ the smallest eigenvalue $\sigma_{\min}(K^\top K) > 0$ is positive.
\item After finitely many iterations $t$ the penalty parameter $\rho$ is sufficiently large and kept fixed such that
\begin{align} \label{eq:penalty_condition}
\frac{L^2}{\rho\sigma_{\min}(K^\top K)} +\frac{m-\rho \sigma_{\min}(K^\top K)}{2} < 0.
\end{align}
\end{itemize}
When the MRF subproblem is solved to global optimality, convergence can be guaranteed by considering a pointwise minimum over exponentially many augmented Lagrangians and applying existing theory \cite{li2015global,hong2016convergence}.
For the general case, however, the theory needs to be extended. 
Our convergence proof borrows arguments from \cite{li2015global,hong2016convergence}, where the convergence of ADMM in the nonconvex setting is shown via a monotonic decrease of the augmented Lagrangian. 
In our case, for a sufficiently large penalty parameter $\rho$, we can achieve a monotonic decrease of the ``discrete-continuous'' augmented Lagrangian \eqref{eq:dc_augm_lagr}, even if the MRF subproblem \eqref{eq:subproblem_mrf} is not solved globally optimal. This allows us to stop exact MRF solvers early or to apply heuristic solvers if computing global optima is intractable.

For the complete proofs of all the theoretical results, presented in this section, cf. the Appendix~\ref{sec:app:theory}.

\begin{lem} \label{lem:sufficient_descent}
Let $K \in \R^{|\mathcal{V}| \times |\mathcal{V}|}$ be surjective and $\delta \geq 0$. For $\rho$ meeting condition \eqref{eq:penalty_condition} we have that
\begin{enumerate}
\item The ``discrete-continuous'' augmented Lagrangian \eqref{eq:dc_augm_lagr} decreases monotonically with the iterates $(\alpha^{t}, \beta^{t}, \lambda^{t}, y^{t})$:
\begin{multline} \label{eq:sufficient_descent}
\mathfrak{L}_\rho(\alpha^{t+1}, \beta^{t+1}, \lambda^{t+1}, y^{t+1}) - \mathfrak{L}_\rho(\alpha^{t}, \beta^{t}, \lambda^{t}, y^{t}) \\
\leq \left(\tfrac{L^2}{\rho\sigma_{\min}(K^\top K)} +\tfrac{m-\rho \sigma_{\min}(K^\top K)}{2} \right) \|\alpha^{t+1} - \alpha^t\|_F^2 \\
-\delta \llbracket y^{t+1} \neq y^t \rrbracket ,
\end{multline}
where $\llbracket \cdot \rrbracket$ denotes the Iverson bracket.
\item $\{\mathfrak{L}_\rho(\alpha^{t+1}, \beta^{t+1}, \lambda^{t+1}, y^{t+1})\}_{t\in \N}$ is lower bounded.
\item $\{\mathfrak{L}_\rho(\alpha^{t+1}, \beta^{t+1}, \lambda^{t+1}, y^{t+1})\}_{t\in \N}$ converges.
\end{enumerate}
\end{lem}

We are now able to guarantee that feasibility is achieved in the limit. This is in contrast to a dual-decomposition approach \cite{komodakis2007mrf,pmlr-v31-zach13a,swoboda-2017} or a Gauss-Seidel quadratic penalty method (with finite penalty parameter $\rho$), used for instance in \cite{steinbrucker2009large,Kim_2017_ICCV}, where a violation of the consensus constraint remains in the limit. Moreover, if $\delta > 0$ is chosen strictly positive, then the discrete variable is guaranteed to converge, i.e.\ for $T$ sufficiently large, we have $y^{t+1}=y^t$ for all $t>T$.
\begin{lem} \label{lem:feasibility}
Let $\{(\alpha^{t}, \beta^{t}, \lambda^{t}, y^{t})\}_{t\in \N}$ be the iterates produced by Alg.~\ref{alg:dc_admm}. Then $\{(\alpha^{t}, \beta^{t}, \lambda^{t}, y^{t})\}_{t\in \N}$ is a bounded sequence. Furthermore, for $t\to \infty$ the distance of two consecutive continuous iterates vanishes, and feasibility is achieved in the limit:
\begin{align}
\|\alpha^{t+1} - \alpha^t\|_F \to 0, \\
\|\beta^{t+1} - \beta^t\|_F \to 0, \\
\|\lambda^{t+1} - \lambda^t\|_F \to 0, \\
\|K \alpha^{t+1} - \beta^{t+1}\|_F \to 0.
\end{align}
Finally, if $\delta> 0$ is chosen strictly positive, then there exists some $T \in \N$ such $y^{t+1}=y^t$ for all $t>T$. 
\end{lem}

The limit points of our algorithm correspond to ``discrete-continuous'' critical points of the augmented Lagrangian. 
\begin{defi}[``Discrete-continuous'' critical point]
We call $(\alpha^*, \beta^*, \lambda^*, y^*)$ a \emph{``discrete-continuous'' critical point} of the ``discrete-continuous'' augmented Lagrangian \eqref{eq:dc_augm_lagr} if it satisfies
\begin{align}
&0 \in \partial (\ell(y_i^*;\cdot ))(\beta_i^*) - \lambda_i^*,&& \forall\; i \in \mathcal{V}  \label{eq:critical_point_primal_1}\\
&0 \in \partial g(\alpha^*) + K^\top \lambda^* \label{eq:critical_point_primal_2}\\
&K\alpha^* = \beta^* \label{eq:critical_point_feasibility},
\end{align}
for $y^*$ with $E_C(y^*) < \infty$ for all $C \in \mathcal{C}$.
Here, $\partial f(x)$ denotes the ``limiting'' subdifferential \cite[Definition 8.3]{Rockafellar-Variational-Analysis} of the function $f$ at $x$ with $f(x) < \infty$.
\end{defi}

\begin{prop} \label{prop:convergence_nonconvex}
Let $\delta \geq 0$. Then any limit point $(\alpha^*, \beta^*, \lambda^*, y^*)$ of the sequence $\{(\alpha^{t}, \beta^{t}, \lambda^{t}, y^{t})\}_{t\in \N}$  is a ``discrete-continuous'' critical point.
\end{prop}

Finally, under convexity of $f$ and $\ell(y_i; \cdot)$, for all $y_i \in \mathcal{L}$ and strictly positive $\delta > 0$, we can guarantee that the sequence of iterates produced by Alg.~\ref{alg:dc_admm} globally converges to a point $(\alpha^*, \beta^*, \lambda^*, y^*)$ which has the following property: $\alpha^*$ is the global optimum of the supervised learning problem w.r.t.\ the estimated training labels $y^*$:
\begin{align} \label{eq:convex_supervised_training}
\alpha^* = \argmin_\alpha \sum_{i \in \mathcal{V}} \ell(y_i^*; K_i \alpha) + f(\alpha).
\end{align}
\begin{prop} \label{prop:convergence_convex}
Let $\ell(y_i; \cdot)$ and $g$ be proper, convex and lower-semicontinuous and let $\delta > 0$. Then the sequence $\{(\alpha^{t}, \beta^{t}, \lambda^{t}, y^{t})\}_{t\in \N}$ produced by Alg.~\ref{alg:dc_admm} converges to a ``discrete-continuous'' critical point $(\alpha^*, \beta^*, \lambda^*, y^*)$ of \eqref{eq:dc_augm_lagr} and $\alpha^*$ solves the problem \eqref{eq:convex_supervised_training} to global optimality.
\end{prop}

\paragraph{Discussion of the assumptions.}
Note that in general the kernel matrix $K$ is not surjective. However, for the strictly positive definite RBF kernel, $K$ is strictly positive definite so that convergence can be achieved for finite $\rho$.
In order to enforce theoretical convergence for general kernels, we may add a small constant to the diagonal of the kernel matrix $K:=K + \gamma I$ that alters the model only slightly. In fact, for the binary SVM, this change is equivalent to replacing the hinge loss with its square.
\begin{figure*}[t!]
\centering
\def\myHeight{5.32em}
\setlength{\tabcolsep}{1pt}
\renewcommand{\arraystretch}{0.5}
\begin{tabular}{lccccc}
 & Frame 5 & Frame 15 & Frame 60 & Frame 68 & Frame 84 \\
\rotatebox{90}{~ \cite{Caelles2017}}  &
\includegraphics[height=\myHeight]{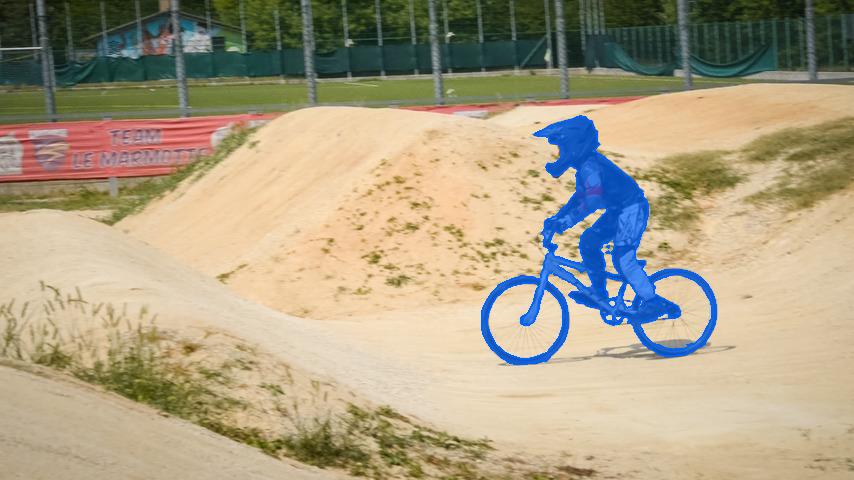} &
\includegraphics[height=\myHeight]{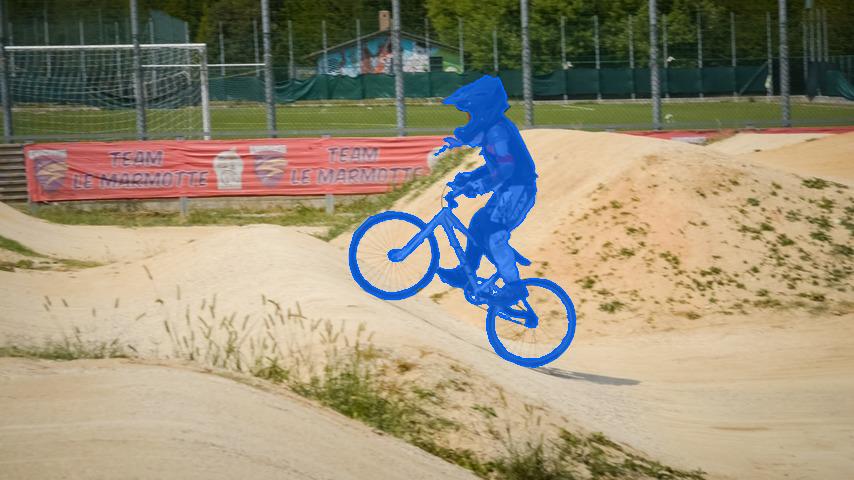} &
\includegraphics[height=\myHeight]{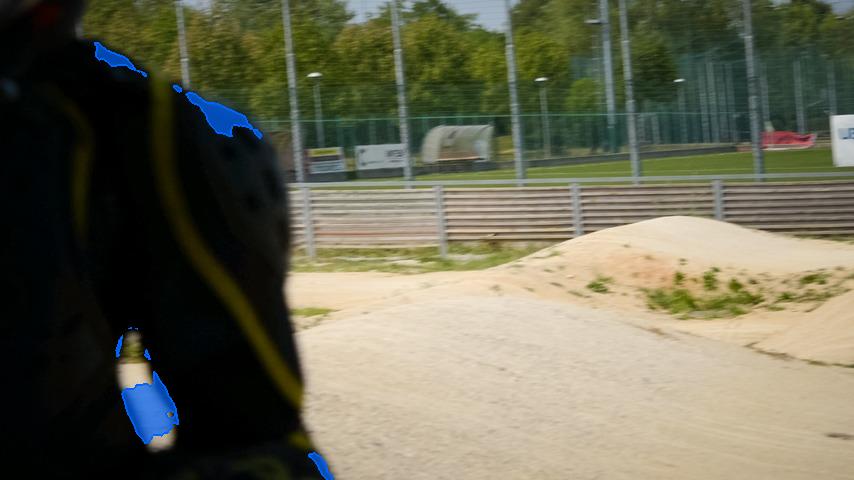} &
\includegraphics[height=\myHeight]{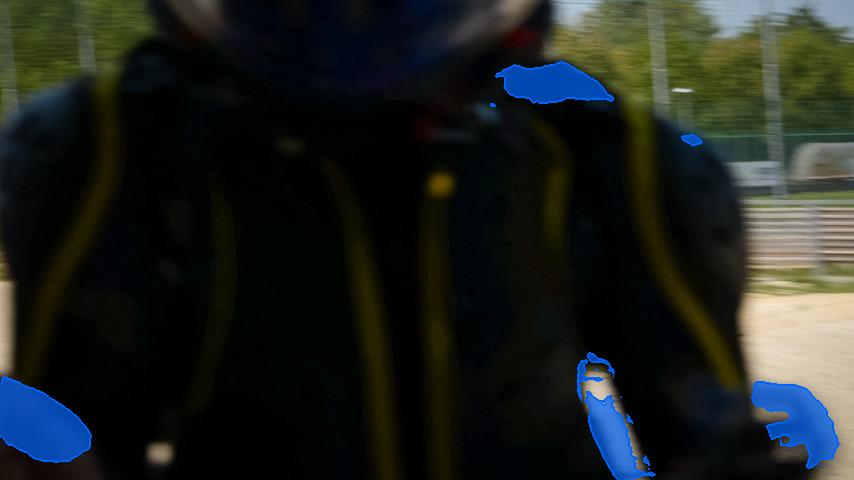} &
\includegraphics[height=\myHeight]{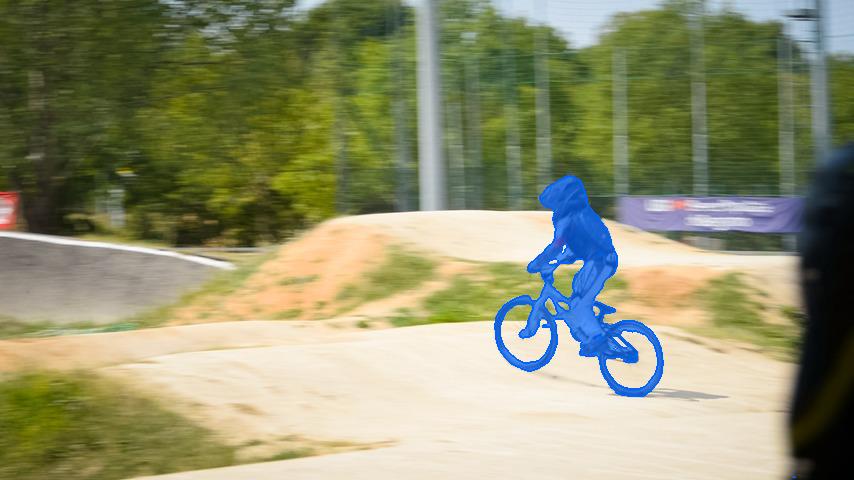} \\
\rotatebox{90}{~ Inductive} &
\includegraphics[height=\myHeight]{bmx-bumps-mrf-fully-supervised-00005.jpg} &
\includegraphics[height=\myHeight]{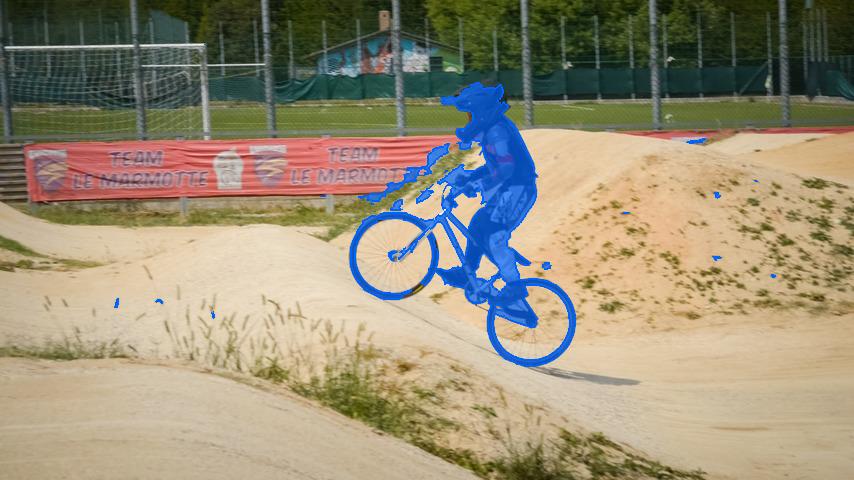} &
\includegraphics[height=\myHeight]{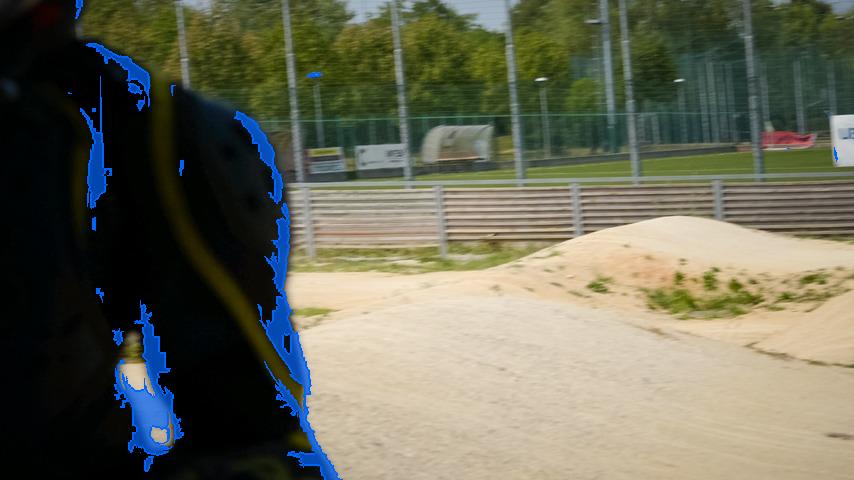} &
\includegraphics[height=\myHeight]{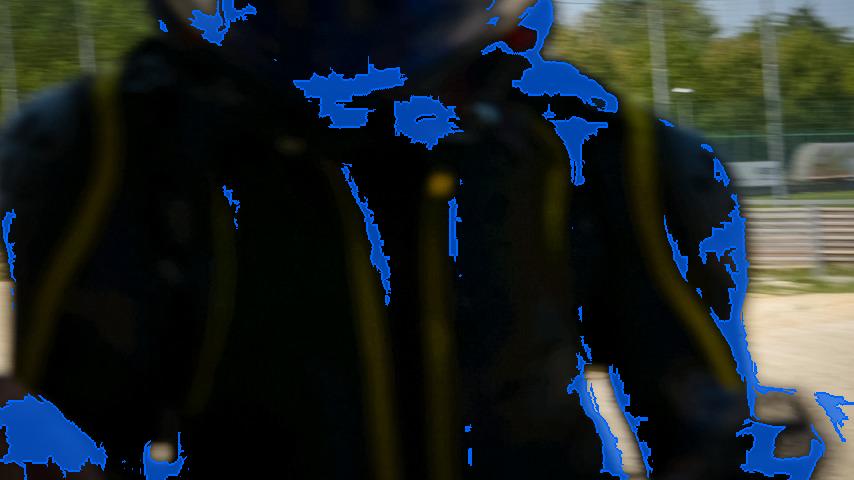} &
\includegraphics[height=\myHeight]{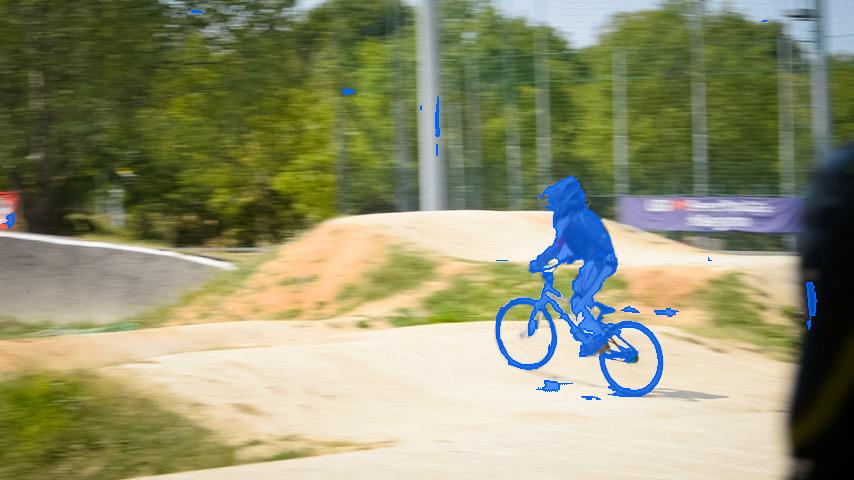} \\
\rotatebox{90}{~ Ours} &
\includegraphics[height=\myHeight]{bmx-bumps-mrf-dc-splitting-00005.jpg} &
\includegraphics[height=\myHeight]{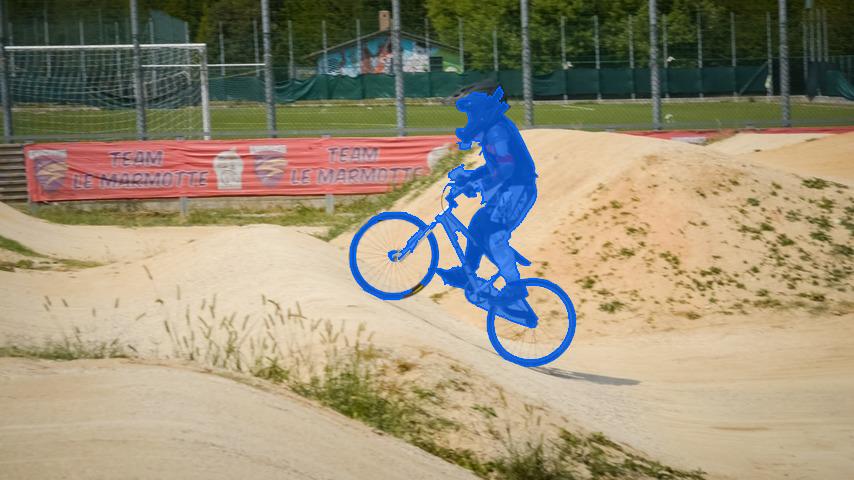} &
\includegraphics[height=\myHeight]{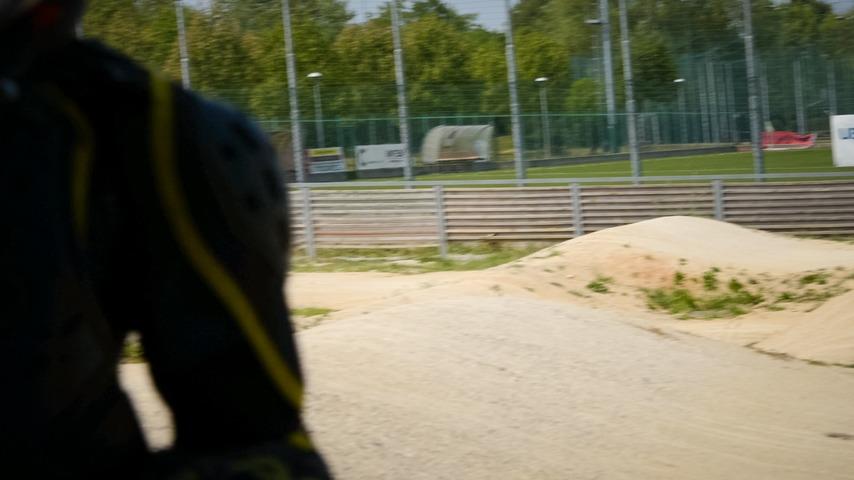} &
\includegraphics[height=\myHeight]{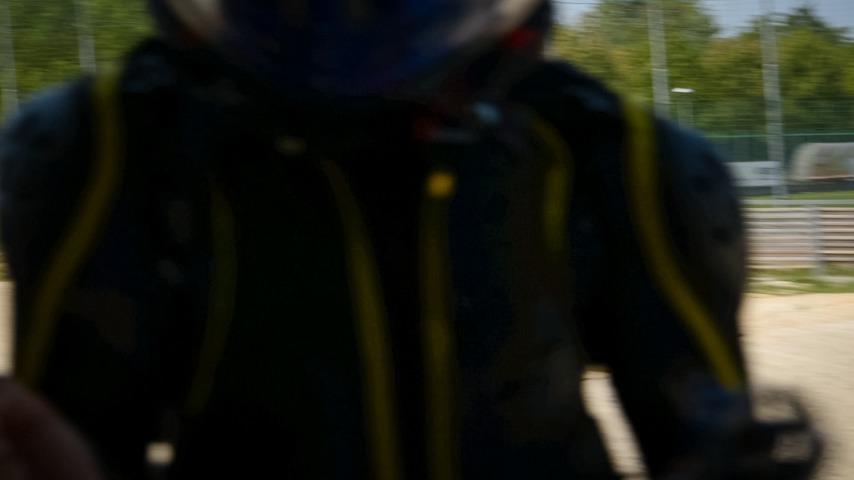} &
\includegraphics[height=\myHeight]{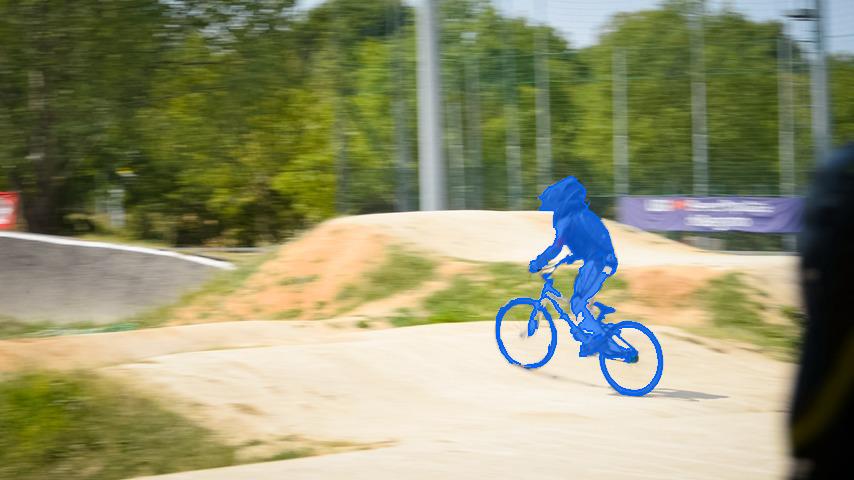}
\end{tabular}
\caption{Exemplary results for video object segmentation on the DAVIS benchmark~\cite{Perazzi2016}. It can be seen that both, the inductive MRF inference approach and OSVOS produce a large number of false positive object pixels for the frames where the object is occluded. \label{fig:videoSegmentation}}
\end{figure*}

\section{Experiments}
In this section, we present the experimental results of our method on several transductive learning tasks. First, we compare our method to an SDP relaxation method for transductive multinomial logistic regression by \cite{DBLP:conf/icml/JoulinB12}.
Second, we use our model and solver for the tasks of object video segmentation as well as image segmentation with user interaction, showing improvements on the false positive rate of object pixels.
{
\begin{table}[t!]
\caption{Comparison with the method of \cite{DBLP:conf/icml/JoulinB12} on the SSL benchmark \cite{Chapelle:2006a}. Reported are the average label-accuracy (in \%) and variance over the splits. Our evaluation suggests that our method performs better for a standard hyperparameter setting except for three out of 20 settings. Moreover, it produces more consistent results, i.e.\ lower variances over the splits. \label{tab:sdp}}
\center
{\scriptsize
\begin{tabular}{ccccccc}
\toprule
& \multicolumn{2}{c}{Linear Kernel} & \multicolumn{2}{c}{RBF Kernel} \\
\cmidrule{2-5}
Dataset & SDP & Ours & SDP & Ours  \\
\midrule
Digit1,10l & 69.27$\pm$27.56     & {\bf82.20$\pm$4.54}  & 53.93$\pm$9.43    & {\bf78.18$\pm$8.43} \\[0.5ex]
USPS,10l & 57.72$\pm$13.73    & {\bf64.58$\pm$3.37}  & 40.10$\pm$11.67  & {\bf48.19$\pm$5.84} \\[0.5ex]
BCI,10l &    50.44$\pm$3.16      & {\bf50.62$\pm$2.08}        & 50.00$\pm$3.06    & {\bf51.67$\pm$0.44} \\[0.5ex]
g241c,10l & 49.88$\pm$38.92    & {\bf55.42$\pm$3.95}  & 62.33$\pm$36.84  & {\bf89.98$\pm$0.32} \\[0.5ex]
g241n,10l & 52.77$\pm$34.37    & {\bf57.61$\pm$4.44}  & 50.13$\pm$0.53    & {\bf51.13$\pm$0.13} \\[0.5ex]
\midrule
Digit1,100l & 75.74$\pm$29.73        & {\bf85.60$\pm$2.91} & {\bf88.65$\pm$0.49}       & 87.61$\pm$3.44 \\[0.5ex]
USPS,100l & 63.44$\pm$9.97         & {\bf72.14$\pm$0.84} & 39.83$\pm$12.63     & {\bf56.54$\pm$3.31}        \\[0.5ex]
BCI,100l & 60.58$\pm$6.87             & {\bf65.23$\pm$1.25} & {\bf64.19$\pm$1.23} & 62.62$\pm$1.00        \\[0.5ex]
g241c,100l & 64.92$\pm$17.47       & {\bf86.31$\pm$0.91}       & 85.63$\pm$0.76        & {\bf89.34$\pm$1.07}  \\[0.5ex]
g241n,100l & {\bf 54.14$\pm$17.13} & 54.11$\pm$0.64      & 52.23$\pm$1.61        & {\bf53.98}$\pm$0.38        \\[0.5ex]
\bottomrule
\end{tabular}
}
\label{tab:bach_sdp}
\end{table}}
\subsection{Comparison with SDP relaxation for transductive learning}
In this experiment, we consider the standard SSL benchmark \cite{Chapelle:2006a} for a comparison with the SDP relaxation method for transductive multinomial logistic regression by \cite{DBLP:conf/icml/JoulinB12}. The benchmark is a collection of several datasets, with varying feature dimensions and number of classes. Each dataset is provided with 12 splits into $l=10$ or $l=100$
labeled and $N-l$ unlabeled samples. We introduce additional unary energies $E_C$ with $|C|=1$ for all the labeled examples, to constrain their label to be fixed during optimization. While \cite{DBLP:conf/icml/JoulinB12} incorporates an entropy prior on the labeling which favors an equal balance distribution, we introduce a higher order potential $E_C$, with $C=\mathcal{V}$, that restricts the solution to deviate at most 10 percent from the equal balance distribution. We solve the LP-relaxation of the higher-order MRF subproblem \eqref{eq:subproblem_mrf} with the dual-simplex method and round the solution. The baseline results are computed with a MATLAB implementation that is provided by the authors. For these experiments, we use the softmax loss and set the regularization parameter $\nu = 0.05$ for the linear kernel. For the RBF kernel we manually chose the variance parameter $\sigma=0.5477$ and the regularization parameter $\nu=0.0025$. We chose the initial penalty parameter $\rho_0 = 0.001$ and $\tau=1.003$. All values are averaged over 12 different splits. The evaluation in Tab.~\ref{tab:sdp} suggests, that our method performs better for a standard hyperparameter setting except for three out of 20 settings. Moreover, it produces more consistent results, i.e.\ lower variances over the splits, which suggests that our method is more robust towards noise and poorly labeled data.

\subsection{Video object segmentation} \label{sec:video-segmentation}
In this experiment, we evaluated our method on video object segmentation. Here, the task is to segment an object throughout a video, given its mask in the first frame. This problem has been successfully approached by \cite{Caelles2017}, using end-to-end deep learning with fully convolutional neural networks. At test time, their classifier is fine-tuned on the appearance of the object and the background in the first frame and predicts the object pixels of individual later frames. However, this method struggles with drastic appearance changes of the object, which have not been learned in advance. These include pose changes, sharp lighting and background changes or severe occlusions as shown in Fig.~\ref{fig:videoSegmentation}.
\begin{figure}[t!]
\centering
\setlength{\tabcolsep}{1pt}
\renewcommand{\arraystretch}{0.5}
\begin{tabular}{lcc}
& Frame 3 & Frame 39 \\
\rotatebox{90}{~ \cite{Caelles2017}} &
\includegraphics[width=0.22\textwidth]{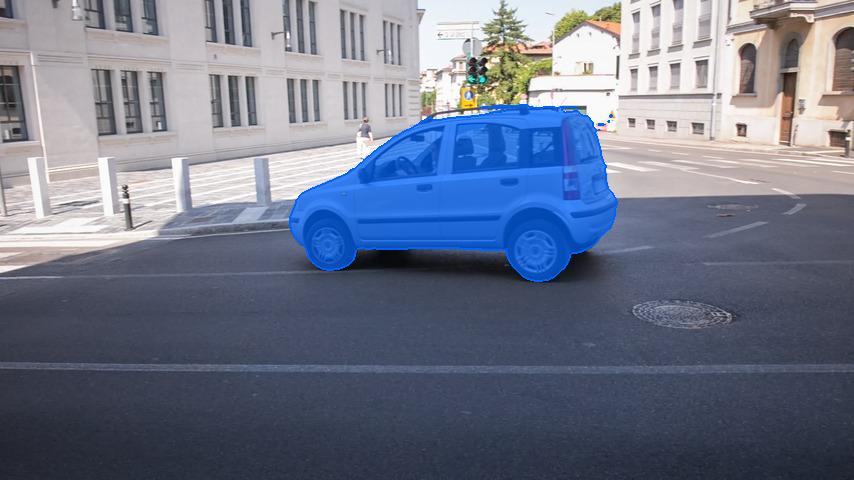} &
\includegraphics[width=0.22\textwidth]{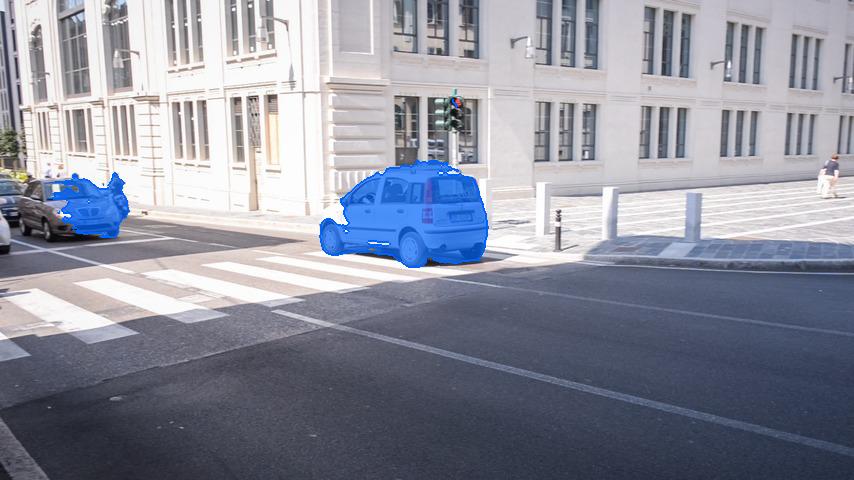} \\
\rotatebox{90}{~ Inductive} &
\includegraphics[width=0.22\textwidth]{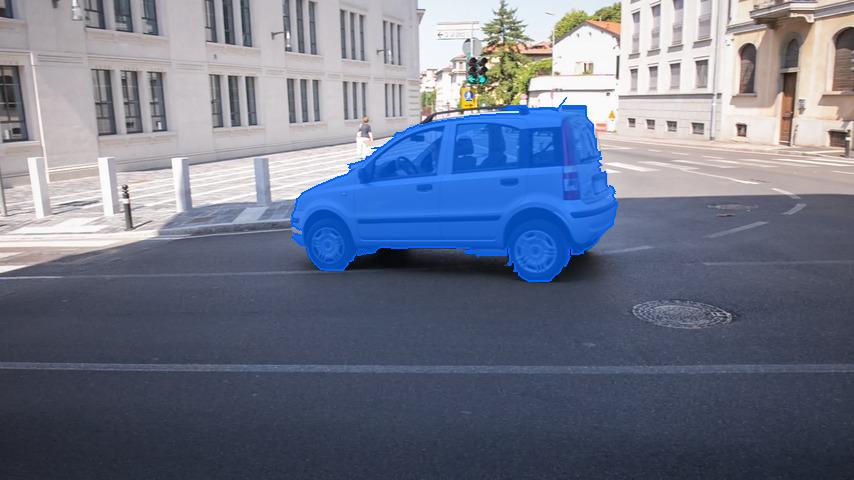} &
\includegraphics[width=0.22\textwidth]{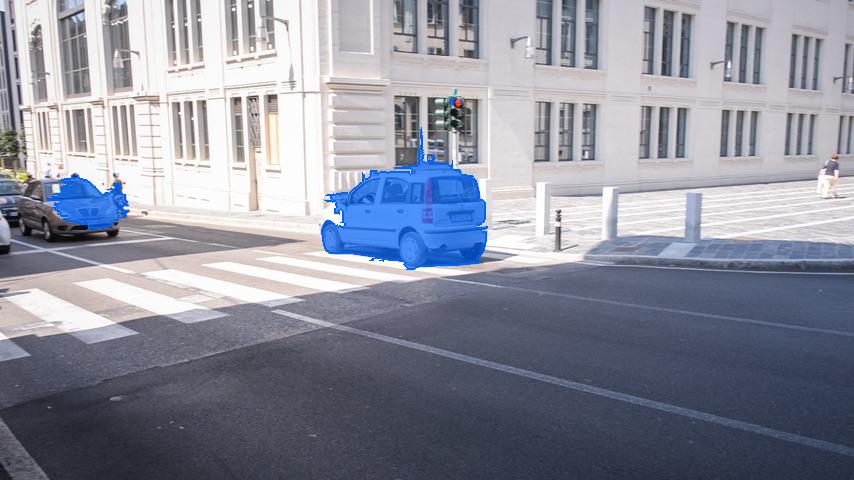} \\
\rotatebox{90}{~ Ours} &
\includegraphics[width=0.22\textwidth]{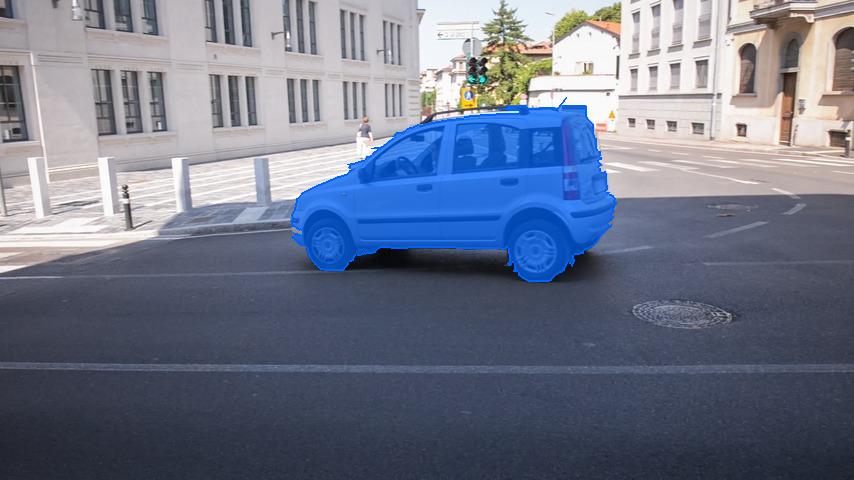} &
\includegraphics[width=0.22\textwidth]{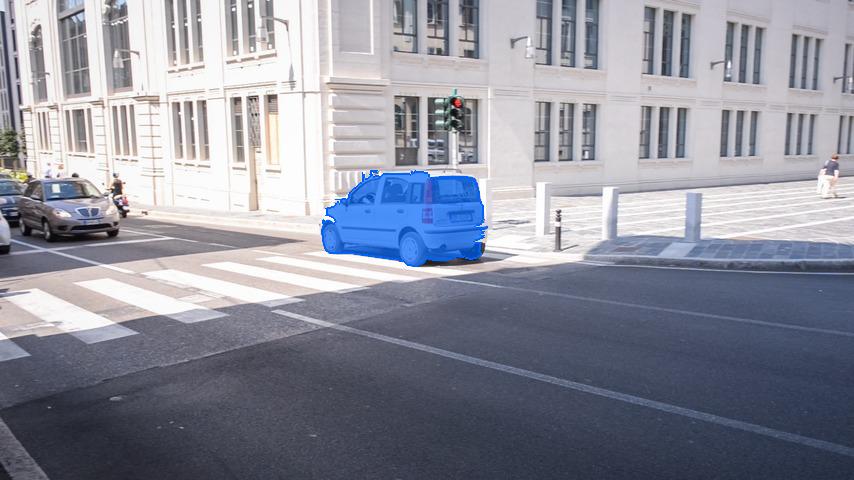} \\
\end{tabular}
\caption{Further results for video object segmentation on the car-shadow sequence from the DAVIS benchmark~\cite{Perazzi2016}. \label{fig:video_segmentation_car-shadow}}
\end{figure}

We propose to use a transductive approach instead. More precisely, we use the pre-trained (not fine-tuned) OSVOS parent network~\cite{Caelles2017} as a deep feature extractor and a MRF model with a variable classifier in the form of \eqref{eq:bilevel_model}. We use a simple linear kernel SVM in our model, as the extracted deep features are almost linearly separable. Further, we introduce unary indicator energies $E_i$ to fix the labels of the user-annotated pixels in the first frame and pairwise energies $E_{ij}$ for adjacent pixels in any frame to favor spatially smooth solutions. Similar to \cite{Caelles2017}, we do not use any temporal consistency terms. 
To reduce the number of examples, we apply our method on a superpixel level and extract 6000 super-pixels~\cite{snic_cvpr17} for each frame and apply average pooling over the superpixels. We compare the proposed transductive approach to OSVOS~\cite{Caelles2017} and the classical (inductive) MRF inference approach (where the classifier is learned with the first frame only) on the DAVIS benchmark~\cite{Perazzi2016}. For both the inductive and the transductive approach the used linear kernel SVM model, the higher order energies $E_C$ and the extracted superpixels are the same.
The results are shown in Fig.~\ref{fig:videoSegmentation}. It can be seen that \cite{Caelles2017} works well as long as the appearance of the object and background are sufficiently similar to the first frame (first column). In frames 60 to 68, where the object is occluded, both the inductive MRF inference approach and OSVOS produce a large number of false positive object pixels. In this experiment the intersection-over-union scores (the higher the better) are 0.7087 for our method, 0.6452 for OSVOS and 0.5063 for the inductive approach.
 Similarly in the car-shadow sequence, OSVOS and the inductive approach mask additionally the other car and the motorbike in frame 39 (cf. Fig.~\ref{fig:video_segmentation_car-shadow}). In contrast our method masks the correct car only. Here, the intersection-over-union scores are 0.9262 for OSVOS, 0.9196 for our method and 0.8844 for the inductive approach.



\subsection{Image segmentation with user interaction}
\begin{figure}[t!]
\centering
\setlength{\tabcolsep}{1pt}
\renewcommand{\arraystretch}{0.7}
\begin{tabular}{ccc}
Annotation & \cite{tang2015secrets,tang2016normalized} & Ours \\
\includegraphics[width=0.155\textwidth]{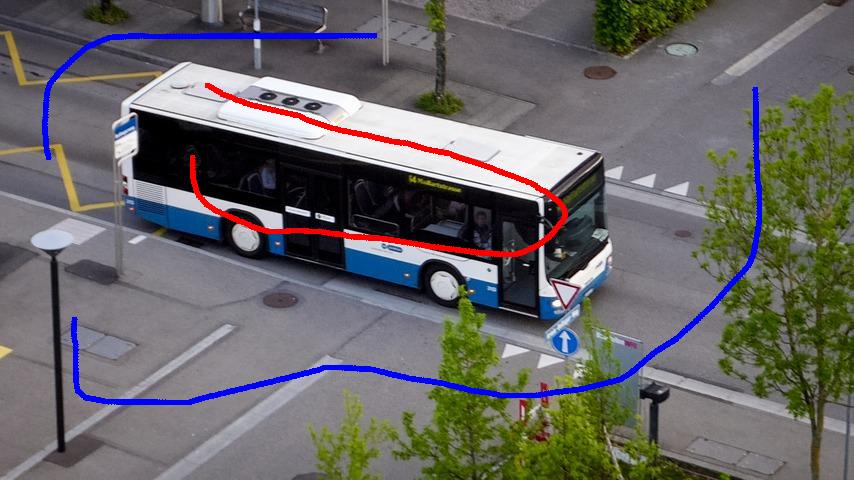} &
\includegraphics[width=0.155\textwidth]{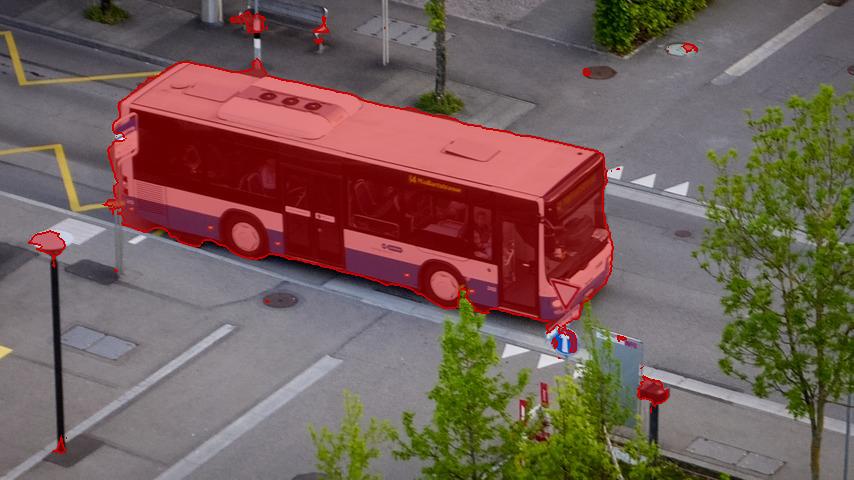} &
\includegraphics[width=0.155\textwidth]{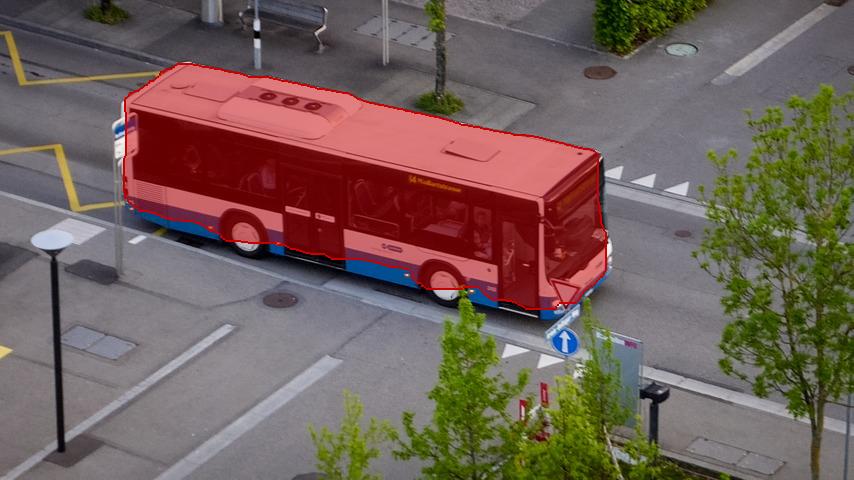} \\
\includegraphics[width=0.155\textwidth]{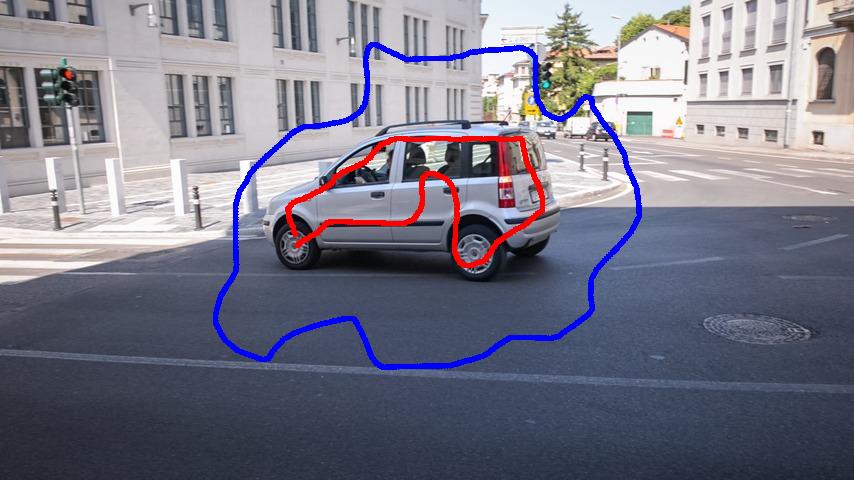} &
\includegraphics[width=0.155\textwidth]{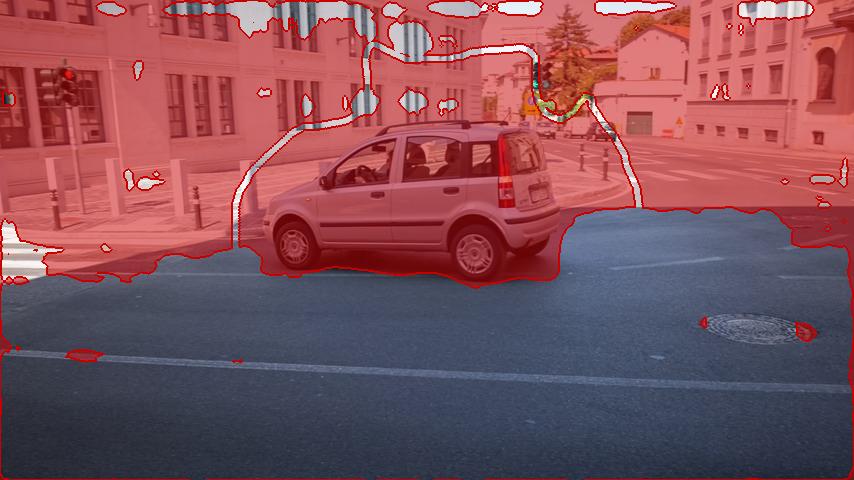} &
\includegraphics[width=0.155\textwidth]{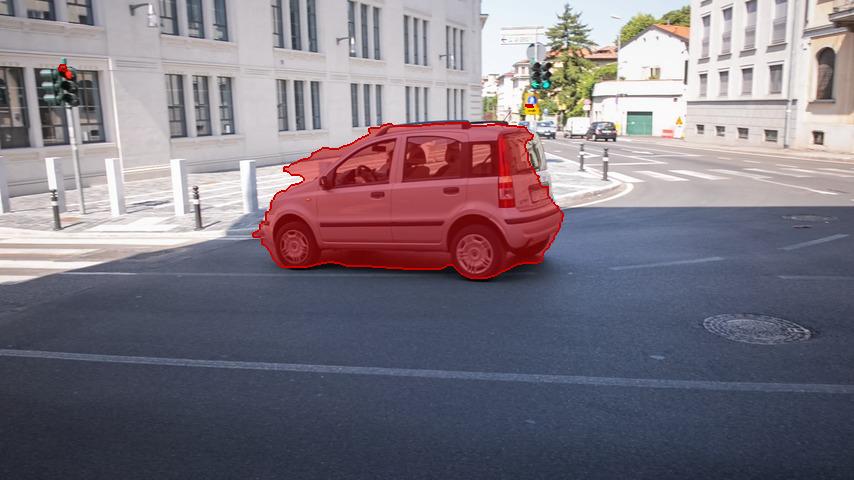} \\
\includegraphics[width=0.155\textwidth]{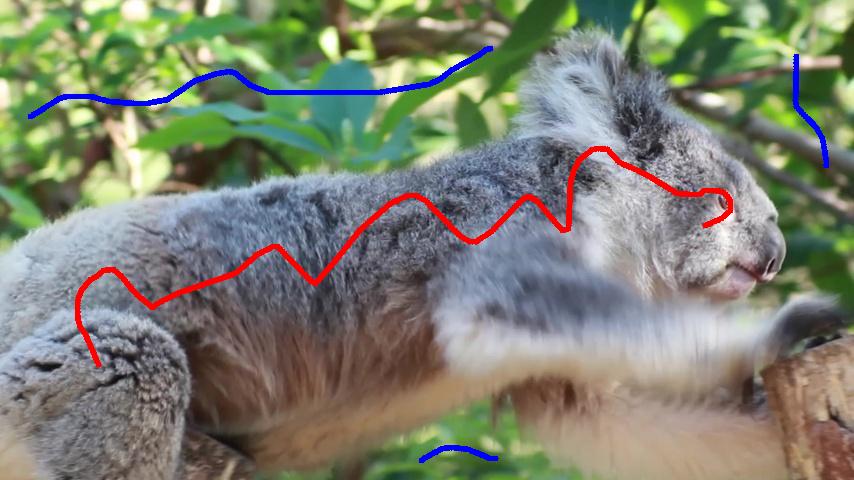} &
\includegraphics[width=0.155\textwidth]{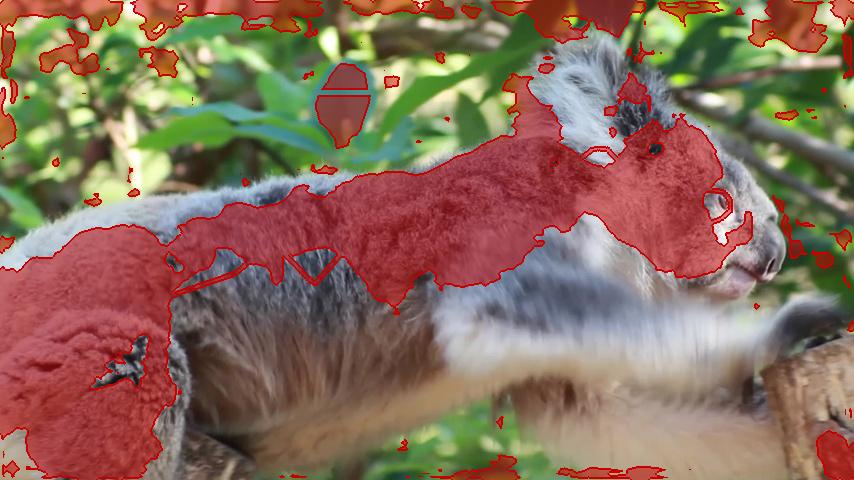} &
\includegraphics[width=0.155\textwidth]{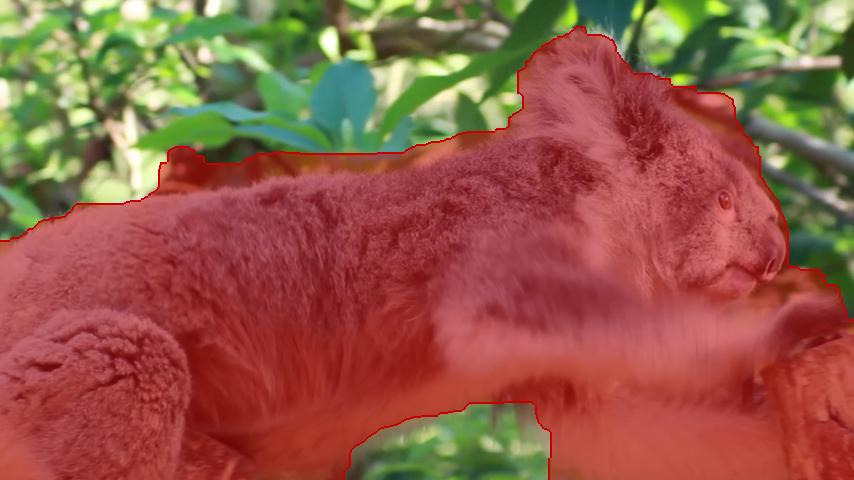}
\end{tabular}
\caption{Exemplary results for interactive binary image segmentation with deep features. {\bf Left}: Input images along with user scribbles in red for foreground and blue for background. {\bf Middle}: Segmentation results (red masks) obtained from $k$-means. {\bf Right}: Segmentation results (red masks) obtained with the proposed method. \label{fig:userInteraction}}
\end{figure}

We evaluated our method on the task of interactive foreground-background segmentation with deep features. Like in the previous experiment we used OSVOS as a deep feature extractor. On this task we compare our method to the Chan-Vese kernel $k$-means approach proposed in \cite{tang2015secrets,tang2016normalized} as a baseline method. Since the features are almost linearly separable, we use a simple linear kernel for both our model and the baseline model.
As it is shown in Fig.~\ref{fig:userInteraction}, the $k$-means approach often fails to find a good cluster-center assignment, despite of strong supervision (provided in the form of user-scribbles) and richness of the features. This is due to the fact that deep high dimensional features are in general not $k$-means friendly \cite{YangFSH17}, which means further pre-processing or a $k$-means suited kernel would be required. In contrast, our method provides a reasonable result for all cases, without the need for feature-pre-processing or kernel-parameter tuning.

\section{Conclusion}
We considered the joint solution of MAP-inference in MRFs and parameter learning, which can be viewed as a transductive inference problem. To solve this task, we proposed a novel algorithm that jointly optimizes over the discrete label variables and the continuous model parameters. The proposed method is related to classical ADMM from continuous optimization and admits a convergence proof under suitable assumptions even though the objective function is discrete-continuous and nonconvex. Our algorithm makes use of a decoupling of the problem into purely discrete and purely continuous subproblems and can be implemented in a distributed fashion. 
We evaluated our approach in several experiments including video object segmentation and interactive image segmentation. Our results suggest that the proposed optimization method performs favorable compared to alternating optimization (as in $k$-means) and convex relaxations. In particular, this indicates that the presented method also serves as an alternative approach to optimization problems arising in semi-supervised or transductive learning, e.g., in the case of SVMs. Furthermore, the visual results show that the transductive inference model is able to reduce the hallucination of false object pixels in image and video segmentation tasks.

\newpage

\appendix 
\section{Theoretical Results} \label{sec:app:theory}
In the remainder of this section we make use of the following properties of $L$-smooth functions (known as the descent-lemma) and $m$-semiconvexity \cite{ArtinaFS13,moellenhoff-siims-15}, which are standard results and therefore stated without proof:
\begin{lem}
Let $f : \R^{k \times n} \to \R$ be continuously differentiable and let $x,y \in \R^{k \times n}$.
\begin{itemize}
  \item If $f$ is $L$-smooth (meaning that $\nabla f$ is Lipschitz continuous with modulus $L$), then
\begin{align}
    f(y) \leq f(x) + \langle \nabla f(x), y-x \rangle + \frac{L}{2} \|x-y\|_F^2.
\end{align}
  \item If $f$ is $m$-semiconvex (meaning that $f + \frac{m}{2} \norm{\cdot}_F^2$ is convex), then
\begin{align}
    f(y) \geq f(x) + \langle \nabla f(x), y-x\rangle - \frac{m}{2}\|x-y\|_F^2.
\end{align}
\end{itemize}
\end{lem}

For showing convergence we make the following assumptions on our problem:
\begin{itemize}
\item The function $f$ is $L$-smooth, $m$-semiconvex and lower-bounded.
\item For all $y_i \in \mathcal{L}$, $\ell(y_i; \cdot)$ is lower-bounded.
\item The kernel matrix $K \in \R^{|\mathcal{V}| \times |\mathcal{V}|}$ is surjective, i.e.\ the smallest eigenvalue $\sigma_{\min}(K^\top K) > 0$ is positive.
\item After finitely many iterations the penalty parameter $\rho$ is sufficiently large and kept fixed such that condition~\eqref{eq:penalty_condition} holds.
\end{itemize}

\subsection{Proof of Lemma \ref{lem:sufficient_descent}}
In \cite{li2015global,hong2016convergence}, to show convergence of nonconvex ADMM, a monotonic decrease of the augmented Lagrangian is guaranteed. Following a similar line of argument, we show that the ``discrete-continuous'' augmented Lagrangian~\eqref{eq:dc_augm_lagr} monotonically decreases with the iterates. Whereas its value decreases with the primal and discrete variable updates, the dual update yields a positive contribution to the overall estimate. Yet, for $\rho>0$ chosen large enough, $K$ surjective and $f$ being $L$-smooth, this ascent can be dominated by a sufficiently large descent in the primal block $\alpha$, updated last.

We need the following notation. Let $B_{:,y^{t}}^{t+1}$ denote the matrix whose $i$-th row is given by $B^{t+1}_{i,y^t_i}$. In particular, by definition of the $\beta$ update, this means $\beta^{t+1} = B^{t+1}_{:,y^{t+1}}$.

\begin{proof}
We rewrite the difference of two consecutive ``discrete-continuous'' augmented Lagrangians as
\begin{align*}
&\mathfrak{L}_\rho(\alpha^{t+1}, \beta^{t+1}, \lambda^{t+1}, y^{t+1}) - \mathfrak{L}_\rho(\alpha^{t}, B_{:,y^{t}}^{t}, \lambda^{t}, y^{t}) \\
&\;= \mathfrak{L}_\rho(\alpha^{t}, B_{:, y^{t}}^{t+1}, \lambda^{t}, y^{t}) - \mathfrak{L}_\rho(\alpha^{t}, B_{:,y^{t}}^{t}, \lambda^{t}, y^{t}) \\
&\quad +  \mathfrak{L}_\rho(\alpha^{t}, \beta^{t+1}, \lambda^{t}, y^{t+1}) - \mathfrak{L}_\rho(\alpha^{t}, B_{:,y^{t}}^{t+1}, \lambda^{t}, y^{t}) \\
&\quad+  \mathfrak{L}_\rho(\alpha^{t+1}, \beta^{t+1}, \lambda^{t}, y^{t+1}) - \mathfrak{L}_\rho(\alpha^{t}, \beta^{t+1}, \lambda^{t}, y^{t+1}) \\
&\quad +  \mathfrak{L}_\rho(\alpha^{t+1}, \beta^{t+1}, \lambda^{t+1}, y^{t+1})\\
&\quad - \mathfrak{L}_\rho(\alpha^{t+1}, \beta^{t+1}, \lambda^{t}, y^{t+1}) 
\end{align*}
We now bound each of the four differences separately:

Since the augmented Lagrangian is separable in $\beta$ and we solve for any $y_i$ a minimization problem in $\beta_{y_i}$ globally optimal we have that
\begin{align}
\mathfrak{L}_\rho(\alpha^{t}, B_{:,y^{t}}^{t+1}, \lambda^{t}, y^{t}) - \mathfrak{L}_\rho(\alpha^{t}, B_{:,y^{t}}^{t}, \lambda^{t}, y^{t}) \leq 0.
\end{align}

A similar estimate holds for the the discrete variable $y^{t+1}$ due to the update in the algorithm:
\begin{equation}
\begin{aligned}
&\mathfrak{L}_\rho(\alpha^{t}, \beta^{t+1}, \lambda^{t}, y^{t+1}) - \mathfrak{L}_\rho(\alpha^{t}, B_{:,y^{t}}^{t+1}, \lambda^{t}, y^{t})\\
&\qquad \leq -\delta\llbracket y^{t+1} \neq y^t\rrbracket .
\end{aligned}
\end{equation}
Now we devise a bound for the third term given by
\begin{align*}
&\mathfrak{L}_\rho(\alpha^{t+1}, \beta^{t+1}, \lambda^{t}, y^{t+1}) - \mathfrak{L}_\rho(\alpha^{t}, \beta^{t+1}, \lambda^{t}, y^{t+1}) \\
&=f(\alpha^{t+1}) - f(\alpha^t) + \langle K \alpha^{t+1}-K\alpha^t,\lambda^t \rangle \\
&\qquad + \frac{\rho}{2} \|K\alpha^{t+1} - \beta^{t+1}\|_F^2 - \frac{\rho}{2} \|K\alpha^{t} - \beta^{t+1}\|_F^2.
\end{align*}
We apply the identity $\|a+c\|_F^2 - \|b+c\|_F^2  = -\|b-a\|_F^2 + 2\langle a+c, a-b \rangle$ with $a:=K \alpha^{t+1}$, $b:=K \alpha^{t}$ and $c=-\beta^{t+1}$ and obtain
\begin{align*}
&f(\alpha^{t+1}) - f(\alpha^t)- \frac{\rho}{2} \|K\alpha^{t+1} - K \alpha^t\|_F^2 \\
&\qquad+ \langle K \alpha^{t+1}-K\alpha^t,\lambda^t +\rho (K \alpha^{t+1} - \beta^{t+1})\rangle .
\end{align*}
The optimality condition for the update of the variable $\alpha$ is given as 
\begin{align} \label{eq:optimality_alpha}
0 = \nabla f(\alpha^{t+1}) + K^\top (\rho (K \alpha^{t+1} - \beta^{t+1}) + \lambda^t).
\end{align}
We replace the term $\langle K \alpha^{t+1}-K\alpha^t,\lambda^t +\rho (K \alpha^{t+1} - \beta^{t+1})\rangle =  \langle \alpha^{t+1}-\alpha^t,K^\top(\lambda^t +\rho (K \alpha^{t+1} - \beta^{t+1}))\rangle$
and obtain from the optimality condition of the $\alpha$ update that
\begin{align*}
&f(\alpha^{t+1}) - f(\alpha^t)- \frac{\rho}{2} \|K\alpha^{t+1} - K \alpha^t\|_F^2 \\
&\qquad+ \langle \alpha^{t+1}-\alpha^t,-\nabla f(\alpha^{t+1}) \rangle \\
&\leq f(\alpha^{t+1}) - f(\alpha^t)- \frac{\rho \sigma_{\min}(K^\top K)}{2}\|\alpha^{t+1} - \alpha^t\|_F^2 \\
&\qquad+ \langle \alpha^t-\alpha^{t+1},\nabla f(\alpha^{t+1}) \rangle.
\end{align*}
Moreover, due to the $m$-semiconvexity of the $f$ we know that
\begin{equation*}
\begin{aligned}
	&f(\alpha^t) + \frac{m}{2} \|\alpha^{t+1} - \alpha^{t}\|_F^2 \\
	&\qquad\geq f(\alpha^{t+1}) + \langle \nabla f(\alpha^{t+1}), \alpha^{t} - \alpha^{t+1}\rangle.
\end{aligned}
\end{equation*}
Overall we can bound
\begin{equation}
\begin{aligned}
&\mathfrak{L}_\rho(\alpha^{t+1}, \beta^{t+1}, \lambda^{t}, y^{t+1}) - \mathfrak{L}_\rho(\alpha^{t}, \beta^{t+1}, \lambda^{t}, y^{t+1}) \\
&\qquad\leq \frac{m-\rho \sigma_{\min}(K^\top K)}{2}\|\alpha^{t+1} - \alpha^t\|_F^2.
\end{aligned}
\end{equation}
Since by assumption $K$ is surjective, the smallest eigenvalue of $K^\top K$ is greater than zero: $\sigma_{\min}(K^\top K) > 0$. This means there exists some $\rho > 0$ large enough so that $\frac{m-\rho \sigma_{\min}(K^\top K)}{2} < 0$.

Finally, we estimate the last term:
\begin{align*}
&\mathfrak{L}_\rho(\alpha^{t+1}, \beta^{t+1}, \lambda^{t+1}, y^{t+1})- \mathfrak{L}_\rho(\alpha^{t+1}, \beta^{t+1}, \lambda^{t}, y^{t+1}) \\
&\qquad= \langle K \alpha^{t+1} - \beta^{t+1},\lambda^{t+1}-\lambda^t \rangle = \frac{1}{\rho} \|\lambda^{t+1} - \lambda^t\|_F^2.
\end{align*}
From the update of the dual variable and the optimality condition for the $\alpha$ update \eqref{eq:optimality_alpha} it follows that
\begin{align}
-\nabla f(\alpha^{t+1}) =  K^\top \lambda^{t+1}.
\end{align}
Further, since $f$ is $L$-smooth we know that
\begin{align}
\|\nabla f(\alpha^{t+1}) -\nabla f(\alpha^t) \|_F^2  \leq L^2\|\alpha^{t+1} - \alpha^t\|_F^2.
\end{align}
Overall, we obtain
\begin{align*}
\sigma_{\min}(K^\top K)\|\lambda^{t+1} - \lambda^t\|_F^2 &\leq \| K^\top \lambda^{t+1} - K^\top \lambda^t\|_F^2 \\
&\leq L^2\|\alpha^{t+1} - \alpha^t\|_F^2.
\end{align*}
This gives the bound for the last term:
\begin{align*}
&\mathfrak{L}_\rho(\alpha^{t+1}, \beta^{t+1}, \lambda^{t+1}, y^{t+1})- \mathfrak{L}_\rho(\alpha^{t+1}, \beta^{t+1}, \lambda^{t}, y^{t+1}) \\
&\qquad\leq\frac{L^2}{\rho\sigma_{\min}(K^\top K)} \|\alpha^{t+1} - \alpha^t\|_F^2.
\end{align*}
Then, by merging the four estimates we obtain the desired result.

We proceed showing the lower boundedness of $\{\mathfrak{L}_\rho(\alpha^{t+1}, \beta^{t+1}, \lambda^{t+1}, y^{t+1}) \}_{t\in \N}$.
Since $K$ is surjective, there exists $\alpha'$ such that $K\alpha' = \beta^{t+1}$ and it holds that
\begin{align*}
-\frac{L}{2}\|\alpha^{t+1} - \alpha'\|_F^2 &\geq -\frac{L}{2\sigma_{\min}(K^\top K)}\| K \alpha^{t+1} - K \alpha' \|_F^2.
\end{align*}
Let $\rho > \frac{L}{\sigma_{\min}(K^\top K)}$. Then, since $f$ is $L$-smooth, we have
\begin{align*}
& f( \alpha^{t+1}) + \langle \lambda^{t+1}, K \alpha^{t+1} - \beta^{t+1} \rangle \\
&\qquad + \frac{\rho}{2}\|K \alpha^{t+1} - \beta^{t+1}\|_F^2 \\
&= f( \alpha^{t+1}) + \langle K^\top \lambda^{t+1},  \alpha^{t+1} - \alpha' \rangle \\
&\qquad + \frac{\rho}{2}\|K \alpha^{t+1} - \beta^{t+1}\|_F^2 \\
&= f( \alpha^{t+1}) + \langle \nabla f(\alpha^{t+1}) ,  \alpha'-\alpha^{t+1} \rangle \\
&\qquad + \frac{\rho}{2}\|K \alpha^{t+1} - \beta^{t+1}\|_F^2 \\
&\geq f( \alpha') - \frac{L}{2}\|\alpha^{t+1} - \alpha' \|_F^2 \\
&\qquad + \frac{\rho}{2}\|K \alpha^{t+1} - \beta^{t+1}\|_F^2 \\
&\geq f( \alpha') + \frac{\rho\sigma_{\min}(K^\top K) - L}{2\sigma_{\min}(K^\top K)}\| K \alpha^{t+1} - \beta^{t+1}\|_F^2 \geq f( \alpha').
\end{align*}
Overall, since by assumption $f$ and $\ell(y_i; \cdot)$ are bounded from below (for all $y_i \in \mathcal{L}$), this means
\begin{align*}
\{\mathfrak{L}_\rho(\alpha^{t+1}, \beta^{t+1}, \lambda^{t+1}, y^{t+1}) \}_{t\in \N}
\end{align*}
is bounded from below.

Since $\{\mathfrak{L}_\rho(\alpha^{t+1}, \beta^{t+1}, \lambda^{t+1}, y^{t+1}) \}_{t\in \N}$ is monotonically decreasing and bounded from below, $\{\mathfrak{L}_\rho(\alpha^{t+1}, \beta^{t+1}, \lambda^{t+1}, y^{t+1}) \}_{t\in \N}$ converges. This completes the proof.
\end{proof}

\begin{figure*}[!htb] 
\centering
\begin{subfigure}[b]{0.23\textwidth}
\centering
	\includegraphics[width=\textwidth]{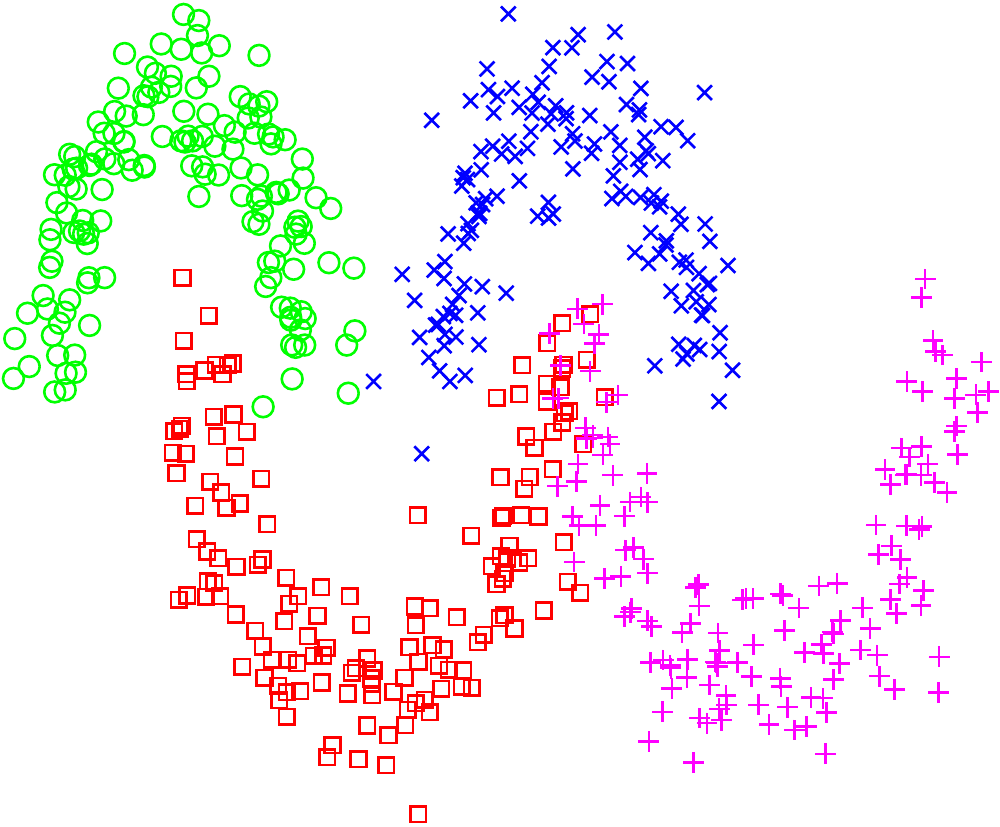}
	\caption{\label{fig:4moons:ground_truth}}
\end{subfigure}
\hfill
\begin{subfigure}[b]{0.23\textwidth}
	\centering
	\includegraphics[width=\textwidth]{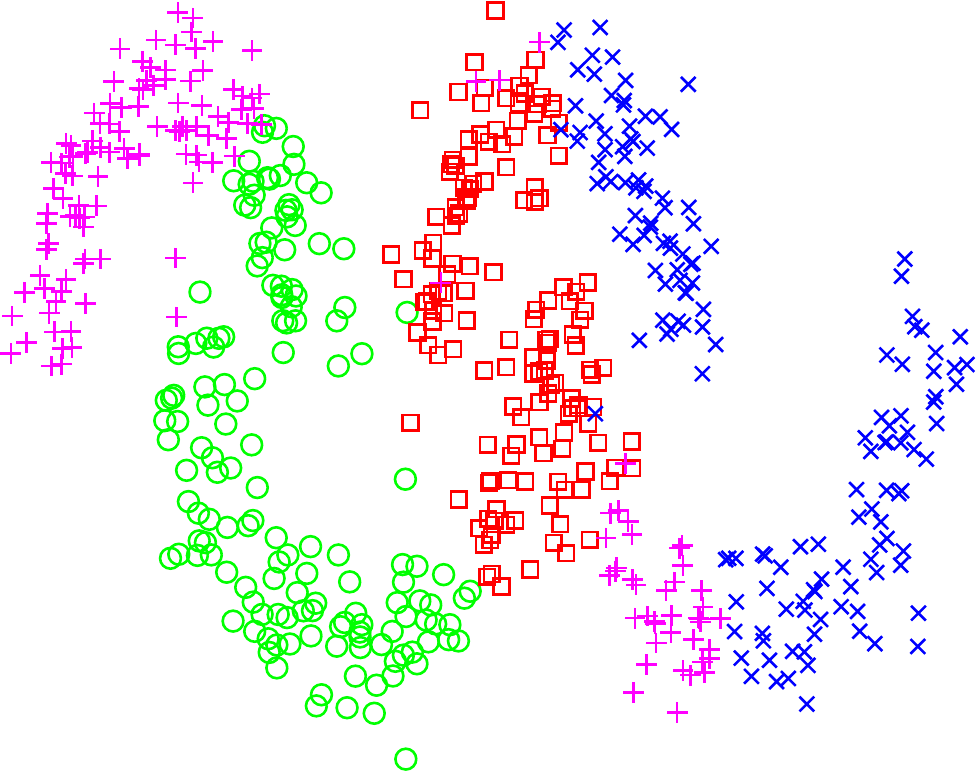}
	\caption{\label{fig:4moons:solution-kkmeans}}
\end{subfigure}
\hfill
\begin{subfigure}[b]{0.23\textwidth}
	\centering
	\includegraphics[width=\textwidth]{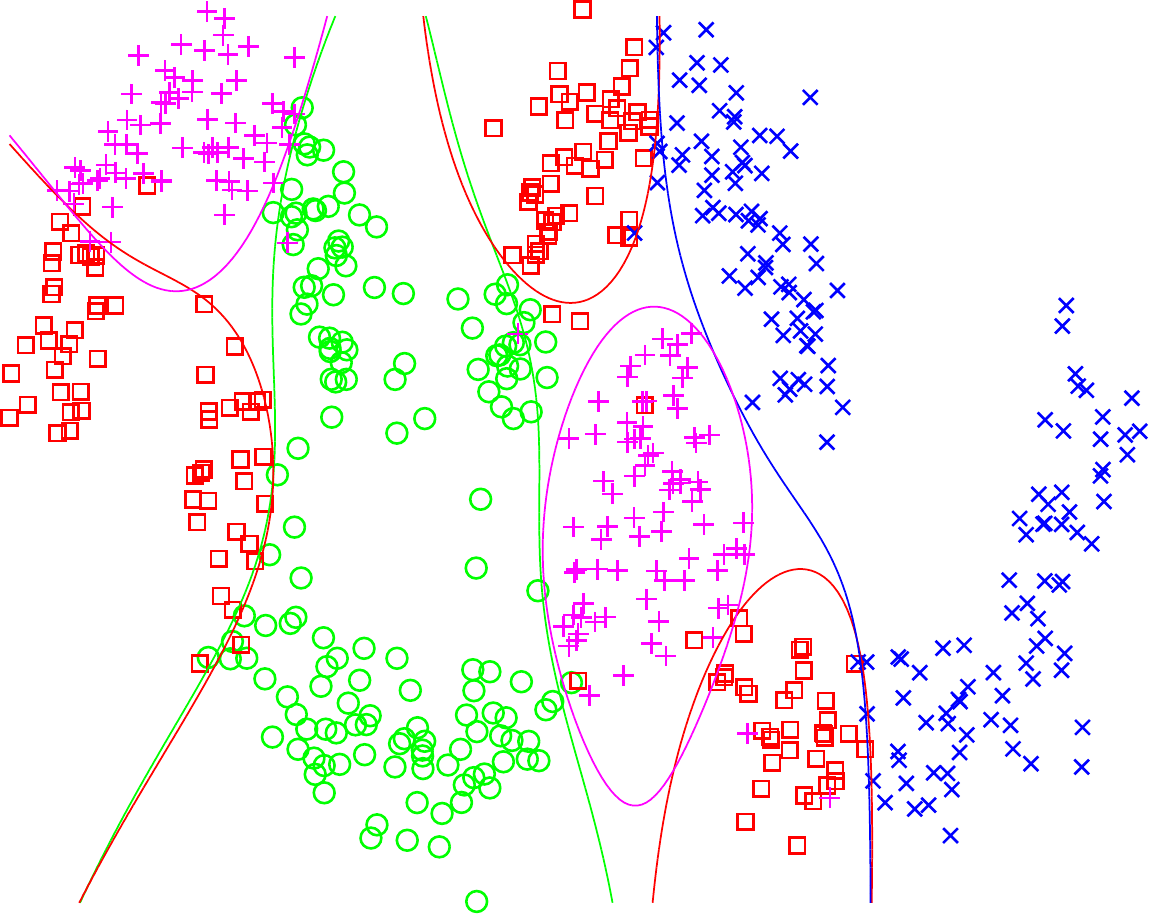}
	\caption{\label{fig:4moons:solution-em}}
\end{subfigure}
\hfill
\begin{subfigure}[b]{0.23\textwidth}
	\centering
	\includegraphics[width=\textwidth]{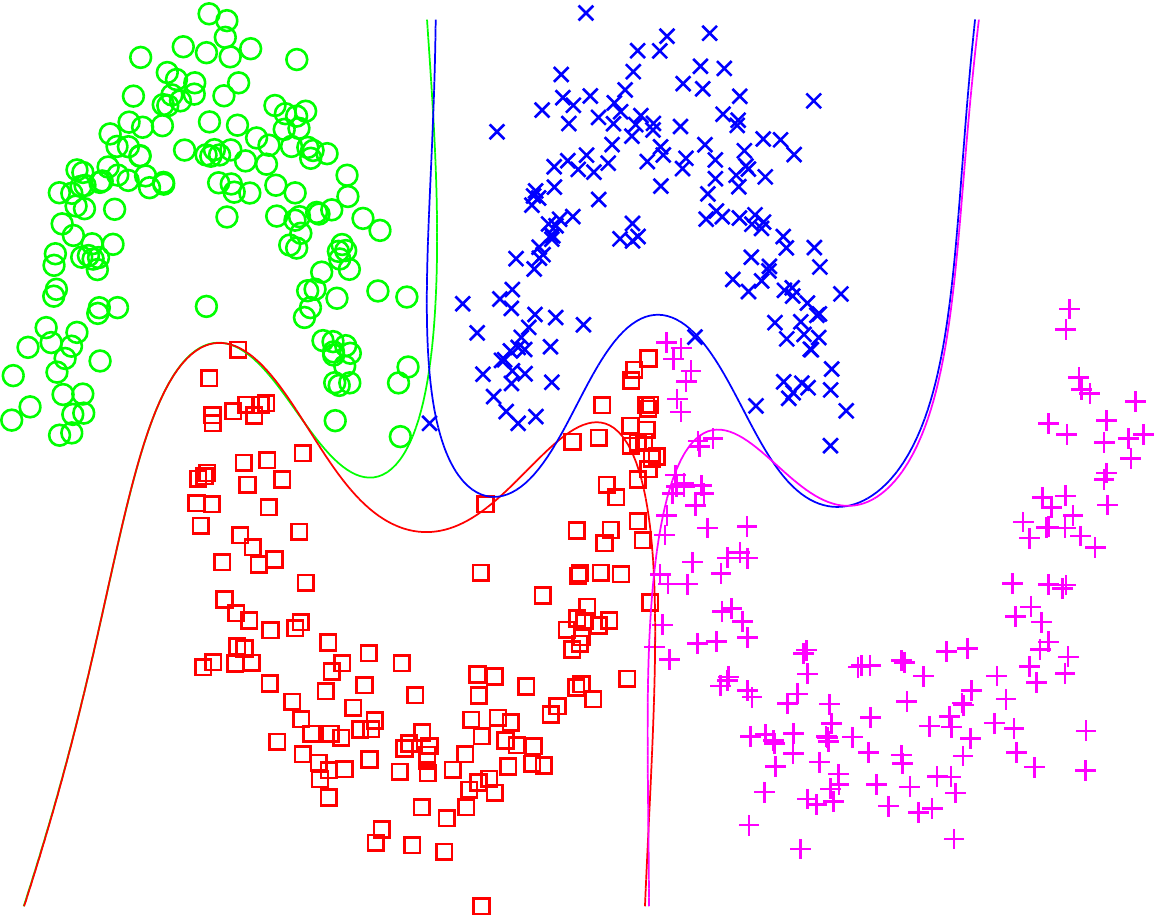}
	\caption{\label{fig:4moons:solution}}
\end{subfigure}
\caption{
Form left to right: Ground-truth, RBF kernel $k$-means, coordinate descent, proposed method.
The label inference errors are 66.6\% for constrained RBF kernel $k$-means, 68.5\% for coordinate descent and 2.5\% for our method.}
\label{fig:4moons}
\end{figure*}
\subsection{Proof of Lemma \ref{lem:feasibility}}
\begin{proof}
We sum over the estimate \eqref{eq:sufficient_descent} which yields
\begin{align*}
-\infty&< \lim_{t\to \infty} \mathfrak{L}_\rho(\alpha^{t}, \beta^{t}, \lambda^{t}, y^{t}) - \mathfrak{L}_\rho(\alpha^{1}, \beta^{1}, \lambda^{1}, y^{1}) \\
&\leq \sum_{t=1}^\infty \left(\tfrac{L^2}{\rho\sigma_{\min}(K^\top K)} +\tfrac{m-\rho \sigma_{\min}(K^\top K)}{2} \right)\|\alpha^{t+1} -\alpha^t\|_F^2 \\
&\quad-\sum_{t=1}^\infty \delta\llbracket y^{t+1} \neq y^t\rrbracket 
\end{align*}
Due to the lowerboundedness, the infinite sums have to converge. This yields that $\|\alpha^{t+1} -\alpha^t\|_F \to 0$. Since $0 \leq \sigma_{\min}(K^\top K)\|\lambda^{t+1} - \lambda^t\|_F^2 \leq \frac{L^2}{\rho\sigma_{\min}(K^\top K)} \|\alpha^{t+1} - \alpha^t\|_F^2$ and $\sigma_{\min}(K^\top K) > 0$ also $\|\lambda^{t+1} - \lambda^t\|_F \to 0$. Since due to the dual update $\lambda^{t+1} - \lambda^t = \rho(K \alpha^{t+1} - \beta^{t+1})$, also
$\|K \alpha^{t+1} - \beta^{t+1}\|_F \to 0$.
Moreover, it holds that
\begin{align*}
\norm{\beta^{t+1} - \beta^t}_F & \leq \norm{\beta^{t+1} - K \alpha^{t+1}}_F + \norm{K\alpha^{t+1} - K \alpha^t}_F \\
& \quad + \norm{K \alpha^t - \beta^t}_F \\
& \leq \norm{K \alpha^{t+1} - \beta^{t+1}}_F \\
& \quad + \norm{K}\norm{\alpha^{t+1} - \alpha^t}_F \\
& \quad + \norm{K \alpha^t - \beta^t}_F \to 0
\end{align*}
for $t \to \infty$.

Finally, suppose that there exists an infinite subsequence $\{t_j\}_{j=1}^\infty \subset \{t\}_{t=1}^\infty$ so that $y^{t_j+1}\neq y^{t_j}$. 
The last sum rewrites as, 
\begin{align*}
\sum_{t=1}^\infty \delta\llbracket y^{t+1} \neq y^t\rrbracket  = \sum_{j=1}^\infty \delta 
\end{align*}
which diverges for $\delta > 0$ positive. This however contradicts the lower boundedness of $\mathfrak{L}_\rho(\alpha^{t}, \beta^{t}, \lambda^{t}, y^{t})$.
\end{proof}

\subsection{Proof of Proposition \ref{prop:convergence_nonconvex}}
\begin{proof}
Let $(\alpha^*, \beta^*, \lambda^*, y^*)$ be a limit point of $\{(\alpha^{t}, \beta^{t}, \lambda^{t}, y^{t})\}_{t\in \N}$, and let $\{t_j\}_{j=1}^\infty \subset \{t\}_{t=1}^\infty$ be the corresponding subsequence of indices. The optimality conditions for the update of the variables $\beta_i$ (for any $i$) and $\alpha$ are given as:
\begin{align}
0 &\in \partial \ell(y_i^{t_j}; \beta_i^{t_j}) - \rho(K_i \alpha^{t_j-1} - \beta_i^{t_j} + \nicefrac{1}{\rho} \lambda_i^{t_j-1}) \\
0 & = \nabla f(\alpha^{t_j}) + \rho K^\top(K\alpha^{t_j} - \beta^{t_j} + \nicefrac{1}{\rho} \lambda^{t_j-1}).
\end{align}
Passing the limit $j\to \infty$ and applying Lemma \ref{lem:feasibility} we arrive at condittions \eqref{eq:critical_point_primal_1}--\eqref{eq:critical_point_feasibility}.
This completes the proof.
\end{proof}
\subsection{Proof of Proposition \ref{prop:convergence_convex}}
\begin{proof}
Let $\delta > 0$. Then, due to Lemma \ref{lem:feasibility} the discrete variable converges, i.e. there is $T>0$ so that for all $t>T$
\begin{align}
y^{t+1}=y^t.
\end{align}
Then, since $f$ and $\ell(y_i; \cdot)$ are convex proper and lsc., after finitely many iterations our scheme Alg. \ref{alg:dc_admm} reduces to convex ADMM and the global convergence is a direct consequence of \cite{gabay1983chapter,eckstein1992douglas,DBLP:journals/ftml/BoydPCPE11}.
This completes the proof.
\end{proof}

\section{Additional Experimental Results}
As a proof of concept we conduct a synthetic experiment with data sampled from 2D moon-shape distributions (600 samples, 4 classes, 150 per class). We sample 25 (possibly overlapping) cliques $C \subset \mathcal{V}$ of cardinality 25 from the set of examples. The synthetic labeling prior in this experiment is given in terms of constraints, that balance the label assignment within each clique. More precisely, it restricts the maximal deviation of the determined labeling from the true labeling to a given bound within each clique $C\in\mathcal{C}$. Mathematically, the higher order energies $E_C$ in the MRF are defined so that $E_C(y_C)= 0$ if $L_C^j \leq |\{i \in C : y_C^i = j \}| \leq U_C^j$, and $\infty$ otherwise. The bounds $L_C^j$ and $U_C^j$ are fixed and chosen a-priori, such that the number of samples $i \in C$ assigned to class $j$ deviates by at most 3 from the true number within clique $C$. This means that we do not provide any exact labels to the algorithm.

The overall task is to infer the correct labels from both, the distribution of the examples in the feature space, and the combinatorial prior encoded within the higher order energies.
Within the algorithm, we solve the LP-relaxation of the higher order MRF-subproblem \eqref{eq:subproblem_mrf} with the dual-simplex method and threshold the solution.
On this task, we compare our method to constrained kernel $k$-means and plain discrete-continuous coordinate descent on \eqref{eq:onelevel_model} with an RBF kernel and an SVM-loss (see Figure \ref{fig:4moons}). Like \cite{tang2015secrets,tang2016normalized,wagstaff2001constrained,Basu-et-al-2008}, we apply $k$-means in the RBF kernel space and solve the E-step w.r.t.\ to \eqref{eq:subproblem_mrf}. It can be seen that both, discrete-continuous coordinate descent on the SVM-based model (Figure \ref{fig:4moons:solution-em}) and constrained kernel $k$-means (Figure \ref{fig:4moons:solution-kkmeans}) get stuck in poor local minima. In contrast, our method is able to infer the correct labels of most examples and finds a reasonable classifier, even for a trivial initialization of the parameters, cf.\ Figure \ref{fig:4moons:solution}. The label errors are 66.6\% for constrained RBF kernel $k$-means, 68.5\% for coordinate descent and 2.5\% for our method.

{\small
\bibliographystyle{ieee}
\bibliography{submission}
}

\end{document}